\documentclass[10pt,twocolumn,letterpaper]{article}

\usepackage{iccv}
\usepackage{times}
\usepackage{epsfig}
\usepackage{graphicx}
\usepackage{amsmath}
\usepackage{amssymb}
\usepackage{multirow}
\usepackage{subcaption}
\usepackage{textgreek}
\usepackage{booktabs}

\usepackage{pdfpages}

\usepackage[breaklinks=true,bookmarks=false]{hyperref}

\iccvfinalcopy 

\def\iccvPaperID{7212} 

\ificcvfinal\fi

\begin{document}
\title{SASMU: boost the performance of generalized recognition model using synthetic face dataset}

\author{Chia-Chun Chung\textsuperscript{1}{\Large\thanks{Equal contribution}}\quad\ Pei-Chun Chang\textsuperscript{2{\normalsize*}}\quad\ Yong-Sheng Chen\textsuperscript{2}\quad\ HaoYuan He\textsuperscript{1}\quad\ Chinson Yeh\textsuperscript{1}\thanks{Corresponding author}\\
\textsuperscript{1}oToBrite Electronics, Inc., Taiwan\\
\textsuperscript{2}Department of Computer Science, National Yang Ming Chiao Tung University, Taiwan\\
{\tt\small zivzhong@otobrite.com, \{pcchang.cs05, yschen\}@nycu.edu.tw, \{michaelhe, charlesyeh\}@otobrite.com}
}

\maketitle
\ificcvfinal\thispagestyle{empty}\fi

\begin{abstract}
Nowadays, deploying a robust face recognition product becomes easy with the development of face recognition techniques for decades.
Not only profile image verification but also the state-of-the-art method can handle the in-the-wild image almost perfectly.
However, the concern of privacy issues raise rapidly since mainstream research results are powered by tons of web-crawled data, which faces the privacy invasion issue.
The community tries to escape this predicament completely by training the face recognition model with synthetic data but faces severe domain gap issues, which still need to access real images and identity labels to fine-tune the model.
In this paper, we propose SASMU, a simple, novel, and effective method for face recognition using a synthetic dataset.
Our proposed method consists of spatial data augmentation (SA) and spectrum mixup (SMU). 
We first analyze the existing synthetic datasets for developing a face recognition system.
Then, we reveal that heavy data augmentation is helpful for boosting performance when using synthetic data.
By analyzing the previous frequency mixup studies, we proposed a novel method for domain generalization. 
Extensive experimental results have demonstrated the effectiveness of SASMU, achieving state-of-the-art performance on several common benchmarks, such as LFW, AgeDB-30, CA-LFW, CFP-FP, and CP-LFW.
\end{abstract}

\section{Introduction}\label{sec:introduction}
With the developments of deep learning techniques, state-of-the-art face recognition methods~\cite{wang2017normface,wang2018cosface,deng2019arcface,sun2020circle,deng2021variational,an2022killing} advance the performance to a great extent, such as over 99.5\% validation accuracy on Labeled Faces in the Wild (LFW)~\cite{huang2008labeled} dataset and 97.70\% TAR@FAR=1e-4 on IJB-C~\cite{maze2018iarpa} dataset. Beyond these successes, researchers also extend the potentials of modern face recognition techniques to some special applications, such as the face with facial mask~\cite{zhu2021masked,deng2021masked} and under the near-infrared light scenario~\cite{kim2019fine_tune_nir,wang2020facial_cyclegan,miyamoto2021joint_nir_vis}. 
However, these methods are mostly trained with web-crawled datasets, such as MS1M~\cite{guo2016ms}, CASIA-Webface~\cite{yi2014learning_casia_webface}, and WebFace260M~\cite{zhu2021webface260m}. Several issues remain challenging, as listed below:
 \begin{itemize}
     \item Privacy issue: It is extremely difficult to collect consents from all enrolled participants, particularly for huge datasets, such as Webface260M~\cite{zhu2021webface260m} containing four million people and over 260 million face images.
     \item Long-tailed distribution: Large differences exist in datasets in terms of the numbers of images, poses, and expressions per person.
     \item Image quality: It is difficult to maintain the same quality level for each image in a large dataset.
     \item Noisy label: Web-crawled image dataset faces the issue of noisy labels when social networks automatically label face images among users and incorrect labeling may occur from time to time.
     \item Lack of attribute annotations: Detailed annotations for facial attributes, such as pose, age, expression, and lighting, are usually not available.
 \end{itemize}
 The most critical challenge is the privacy issue, which we define as whether to use recognizable information or not. To refrain from privacy invasion, unrecognizable noises or random-region masks can be added to face images~\cite{wang2022facemae}, but the risk of leakage of real and distinguishable face images remains high.
 To solve the privacy issue once and for all, using synthetic data to train the face recognition model is a good practice.
 Thanks to the development of the generative model and computer graphics, we could generate realistic images by using computing resources~\cite {deng2020discofacegan,bae2023digiface}.
 However, the domain gap is unavoidable, and the previous works~\cite{qiu2021synface, bae2023digiface} access the real image and label to close the domain gap which still violates privacy-preserving. 
 In this work, our main contributions are summarized as followed:
 \begin{itemize}
     \item Analyzing the impacts and potentials of spatial data augmentation (SA), and showing the analytical results of possible options, such as grayscale, perspective operation, etc.
     \item Applying  the spectrum mixup (\textbf{SMU}) to minimize the synthetic-to-real domain gap without using real face images for training.
     \item Achieving the state-of-the-art performance of face recognition without using any recognizable information.
 \end{itemize}
\section{Related works}\label{sec:related_works}
Recent advancements in face recognition research have focused on improving the accuracy and efficiency of deep learning models. One important research area is to investigate various architectures, such as attention mechanisms~\cite{li2018occlusion_CNN_attention, wang2020hierarchical_attention, wang2021dsa, wang2021aan_face, wang2022cqa_face, lin2021xcos} and multi-task learning~\cite{yin2017multi, huang2021age, neto2021focusface}, to improve model capability in accommodation with variations in facial expressions, poses, and lighting conditions. Most importantly, the design of loss functions~\cite{wang2018cosface, wang2017normface, li2021spherical, deng2019arcface, deng2021variational, sun2020circle, meng2021magface, terhorst2023qmagface, kim2022adaface} and open source large dataset~\cite{yi2014learning_casia_webface, guo2016ms, wang2019rfw, karkkainen2021fairface, zhu2021webface260m} have been drawing much attentions in this research field. Additionally, researchers have been investigating privacy-preserving face recognition techniques to protect individuals' privacy by not storing their raw face images in the system. Instead, they have developed methods to generate privacy-preserving representations for the faces that can be used for person identification.

Recently, researchers have explored the use of synthetic data and applied data augmentation techniques to improve the model generalization capability to compute unseen data. 
Promising results have been reported in reducing bias and improving the recognition accuracy on diverse datasets.
However, there are still certain cases where noticeable gaps exist between real and synthetic images, especially in the frequency domain.
For instance, the artifact on synthetic data can be revealed by using the frequency spectrum analysis and can be recognized by a simple classifier on spectrum \cite{jiang2021focal, durall2020watch}.
Therefore, reducing the domain gap between real and synthetic datasets can lead to more robust and effective recognition systems trained on synthetic data, which can have significant practical applications in areas such as security, surveillance, and biometrics.

To sum up, these recent advancements have the potential to improve the accuracy, efficiency, and fairness of face recognition systems, making them more suitable for real-world applications.

In this section, we first introduce deep face recognition using margin-based softmax loss functions. 
We then explore the performance gap between the models trained on synthetic and real datasets (SynFace and RealFace). 
Lastly, we introduce 1) identity mixup to enlarge the intra-class variations and 2) domain mixup to mitigate the domain gap between synthetic and real face images.

\subsection{Training of face recognition}
Training of face recognition system is to teach a computer vision model to recognize and identify human faces in images or videos. 
The training process involves feeding the model with a large amount of labeled face data, that is, images or videos of people's faces along with corresponding identity labels. 
The model then learns to extract unique features from each face and uses those features to differentiate one's face from another.

During training, parameters of the model are determined through various techniques such as deep learning algorithms, convolutional neural networks (CNNs), and transfer learning to improve its accuracy while reducing errors in identifying faces. 
The accuracy of the face recognition model is evaluated with a separate dataset that is unseen to the model.

The ultimate goal of face recognition training is to develop a robust and reliable system that can accurately recognize and identify faces in real-world scenarios, such as security systems, access control, and biometric authentication.
In the past few years, researchers mainly focus on the loss design for face recognition training.
\subsubsection{Losses for face recognition}
In general, face recognition methods adopt unified (hybrid) loss functions, which combine three losses involving specific constraints on the angles between the weight vectors of different classes, and govern the distance metrics among them.

Let $x_i$ be the input feature vector under class $i$, $y_{i}$ be the ground truth label of class $i$, and $\textit{\textbf{W}}$ be the learnable weight matrix of the loss function with size of $C \times D$, where $C$ is the number of classes, and $D$ is the feature dimension.  
The joint loss function can then be written as:

\begin{equation}
\begin{split}
L & = -\frac{1}{N} \sum_{i=1}^{N} \log \frac{e^{s \cdot \delta}}{e^{s \cdot \delta} + \sum_{j \neq \textit{y}_i} e^{s \cdot \cos(\theta_j)}} \\
\delta & = \cos(m_{1} \cdot \theta_j + m_{2}) - m_{3} \,,
\end{split}
\end{equation}

\noindent where $N$ is the batch size, $m_{i \in {1,2,3}}$ is the margin parameter, $s$ is the scale factor, and $\theta_j$ indicates the angle between the weight $W_j$ and the feature $x_{i}$. In detail, sphereface~\cite{Liu_2017_sphereface}, arcface~\cite{deng2019arcface}, and cosface~\cite{wang2018cosface} have the parameters $(m_1, 0, 0)$, $(1, m_2, 0)$, and $(1, 0, m_3)$,
respectively.
By using this joint loss function, we can train the network to learn discriminative features that are well-separated in the embedding space, while also minimizing the computational overhead of multiple loss functions.


Recent research studies begin to investigate other designs of loss function, such as those considering the image quality effect~\cite{meng2021magface, terhorst2023qmagface, kim2022adaface}. In this paper, we conduct experiments on the SOTA method, Adaface~\cite{kim2022adaface}, which proposed the adaptive margin function by approximating the image quality with feature norms. Their margins can be written as:
\begin{equation}
\begin{split}
m_{1}^{Adaface} & = 1 \\
m_{2}^{Adaface} & = - m_{2} \cdot  \widehat{\| z_{i} \|} \\
m_{3}^{Adaface} & = m_{3} \cdot \widehat{\| z_{i} \|} + m_{3} \\
\widehat{\| z_{i} \|} & = \lfloor \frac{\| x_{i} \| - \mu_z}{\sigma_z / h } \rceil \,,
\end{split}
\end{equation}

\noindent where $\| x_{i} \|$ means the norm of the feature vector $x_{i}$, $h$ is a constant, $\mu_z$ and $\sigma_z$ are the mean and standard deviation of all
$\| x_{i} \|$ within a batch.

\subsection{Privacy-preserving face recognition}
Privacy-preserving face recognition is an emerging area of research. One approach is to use Masked Autoencoders in FaceMAE~\cite{wang2022facemae}, where face privacy and recognition performance are considered simultaneously. Alternatively, learnable privacy budgets in the frequency domain~\cite{ji2022ppfr_learned_budgets} can be used. The other approach is to use differential privacy to convert the original image into a projection on eigenfaces and to add noises for better privacy. Specifically, differential privacy works by adding random noises to the data or query results in a way that guarantees whether the presence or absence of any individual in the dataset does not significantly affect the outcome. This means that an observer cannot determine whether any individual’s data is included in the dataset or not. The amount of noise added is calibrated based on a privacy parameter called epsilon $\epsilon$, which determines the level of privacy protection. A smaller value of ε provides stronger privacy protection but may result in less accurate query results. Differential privacy offers a theoretical guarantee of privacy~\cite{chamikara2020ppfr_diffential_privacy}. 
However, the methods above still need to access real information such as RGB images and related identity information. To refrain from any privacy invasion, we need to 
use synthetic data and avoid any privacy information in the training pipeline. Synface~\cite{qiu2021synface} use the pre-trained GAN~\cite{deng2020discofacegan} to generate massive of synthetic images to decrease the needy of the real image, which only uses 1 over 10 real image~\cite{yi2014learning_casia_webface} for domain mixup in the training pipeline.
\section{Methods}\label{sec:methods}
To improve the accuracy of face recognition network, in this study, we propose a spatial augmentation and spectrum mixup (SASMU) module which operates in the spatial and frequency domains, respectively.
We first describe how the synthetic faces are controlled, rendered, and aligned to prepare the dataset (Section \ref{sec:DA}).
After providing the dataset statistics (Section \ref{sec:DS}), we introduce the proposed spectrum mixup method for minimizing the synthetic-to-real domain gap (Section \ref{sec:SMU}).

\subsection{Dataset statistics}\label{sec:DS}
\begin{figure}[t]
\centering
    \includegraphics[width=\linewidth]{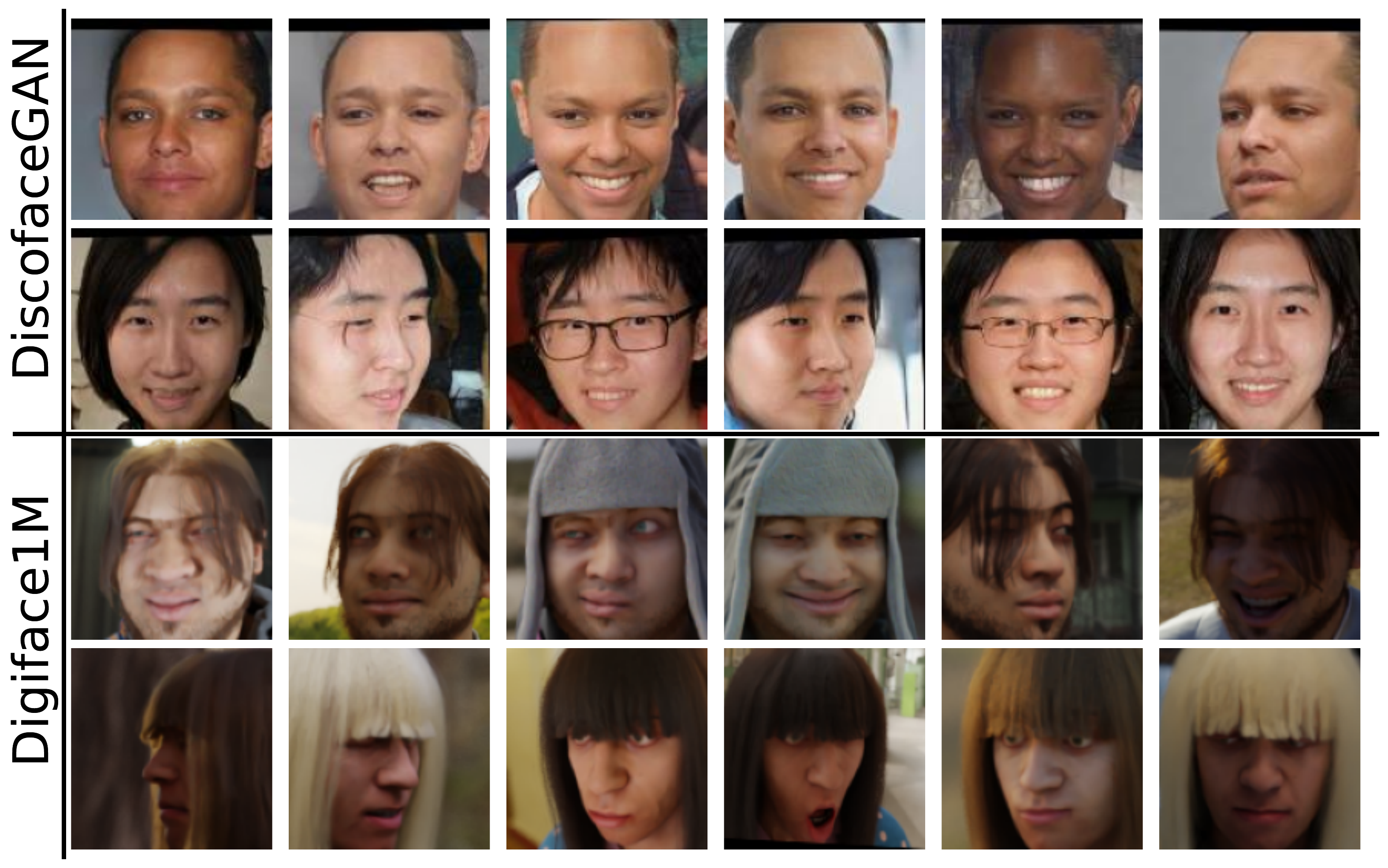}
\caption{Comparison of DiscofaceGAN~\cite{deng2020discofacegan} and Digiface1M~\cite{bae2023digiface} datasets. For each dataset, the images of the first and second rows are samples from two different identities. Obviously, DiscofaceGAN~\cite{deng2020discofacegan} fails to maintain identity consistency among generated face images, whereas Digiface1M~\cite{bae2023digiface} performs great in generating various face images for the same identity.
}
\label{fig:discofacegan_vs_digiface1m}
\end{figure}

In this paper, we conduct our study upon the setting of SOTA face recognition with synthetic image data, Synface~\cite{boutros2023USynthFace} and Digiface-1M~\cite{bae2023digiface}. These two works develop their face recognition model on two different types of datasets, in which the former one use the generated data from GAN and the later one create the images using the traditional simulation and rendering pipeline. Specifically, Synface~\cite{boutros2023USynthFace} uses the pretrained DiscofaceGAN~\cite{deng2020discofacegan} to generate the facial images. DiscoFaceGAN~\cite{deng2020discofacegan} can generate realistic and diverse face images of virtual people with disentangled and controllable features. It uses 3D priors to learn latent representations for identity, expression, pose, and illumination, and then synthesizes face images by imitating a 3D face deformation and rendering process. DiscoFaceGAN~\cite{deng2020discofacegan} can produce high-quality face images with fine-grained control over each feature. Despite the fast synthetic images production, DiscofaceGAN~\cite{deng2020discofacegan} suffers from identity consistency, especially in large poses or severe environment conditions shown in Fig.~\ref{fig:discofacegan_vs_digiface1m}. In contrast, Digiface-1M~\cite{bae2023digiface} leverage the well-developed 3D CG technique to render the image. They first create the synthetic 3D model of a person then follow a predefined instructions to obtain a set of synthesized face images. As shown in Fig.~\ref{fig:discofacegan_vs_digiface1m}, Digiface-1M well produces a variety of face images for the same identity with given context settings. In the following experiment, we also show that the lack of identity consistency decreases the performance of face recognition.

\subsection{Data augmentation}\label{sec:DA}
Data augmentation is widely used in the vision tasks \cite{shorten2019data_augmentation_survey} to extend the amount and variety of training dataset. 
Starting from light augmentation, such as cropping, rescaling, and photometric jittering, to heavy augmentation including but not limited to random affine, random masking, and warping, performance improvements can be achieved by adopting data augmentation in various vision systems.  
In the face recognition research field, however, we surprisingly found that there is almost no additional data augmentation applied in the previous methods.
One intuitive reason might be assocaited with the increasing amount of datasets~\cite{zhu2021webface260m}, which could meet the requirement of scale issues for model training. In practice, we found that only slight data augmentation, for example, the applied probability of grayscale, $p_{grayscale}=0.05$, is beneficial for using Adaface loss~\cite{kim2022adaface} on real image dataset~\cite{yi2014learning_casia_webface}. However, the performance decreases rapidly when $p_{grayscale}$ becomes larger.

In contrast with the diversity of real images, synthetic images are usually monotonous. For example, the face in the real image could be occluded, over-exposed, and distorted. It is hard for GAN model~\cite{deng2020discofacegan} to mimic those artifacts without changing the identities. As for traditional rendering pipeline, the current technique~\cite{bae2023digiface} could generate various face images for the same identity, at the expense of high costs of hardware, software, and computational time. 

In our study, the data augmentation is beneficial for using synthetic data. Taking Synface training method as an example, adding random erase (RE) with probability $p_{RE}=1$ can increase the testing performance on LFW~\cite{huang2008labeled} by more than $1\%$. It is worthwhile to investigate the data augmentation when using synthetic image to train the face recognition model.

We split the data augmentation into two groups, appearance-based and geometry-based methods. While keeping the structure of the face, the appearance-based method only changes the color tonality, such as grayscale, Gaussian noise, blur, salt and pepper, channel shuffle, equalization, and auto contrast. The geometry-based method changes the structure of face images, such as crop, flip, affine, and perspective.



\subsection{The proposed spectrum mixup (SMU)}\label{sec:SMU}
\begin{figure}[t]
\centering
    \includegraphics[clip,trim= 0cm 9.5cm 17cm 2cm,width=0.8\linewidth]{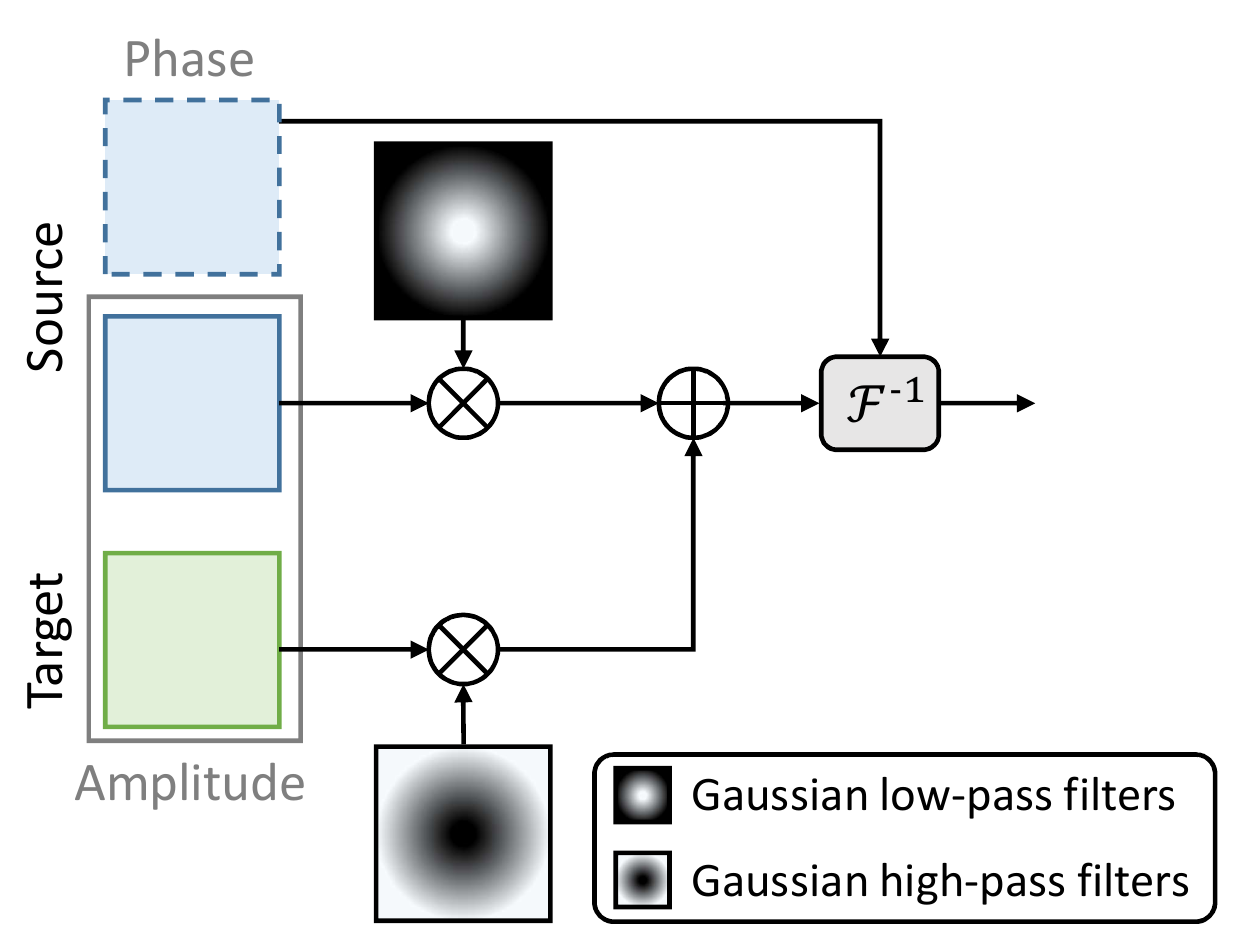}
\caption{The proposed spectrum mixup (SMU) module.}
\label{fig:SMU}
\end{figure}

\begin{figure*}[t]
\centering
    \begin{tabular}{cccc}
        \includegraphics[clip,trim= 0cm 9cm 22.5cm 2cm,width=0.18\textwidth]{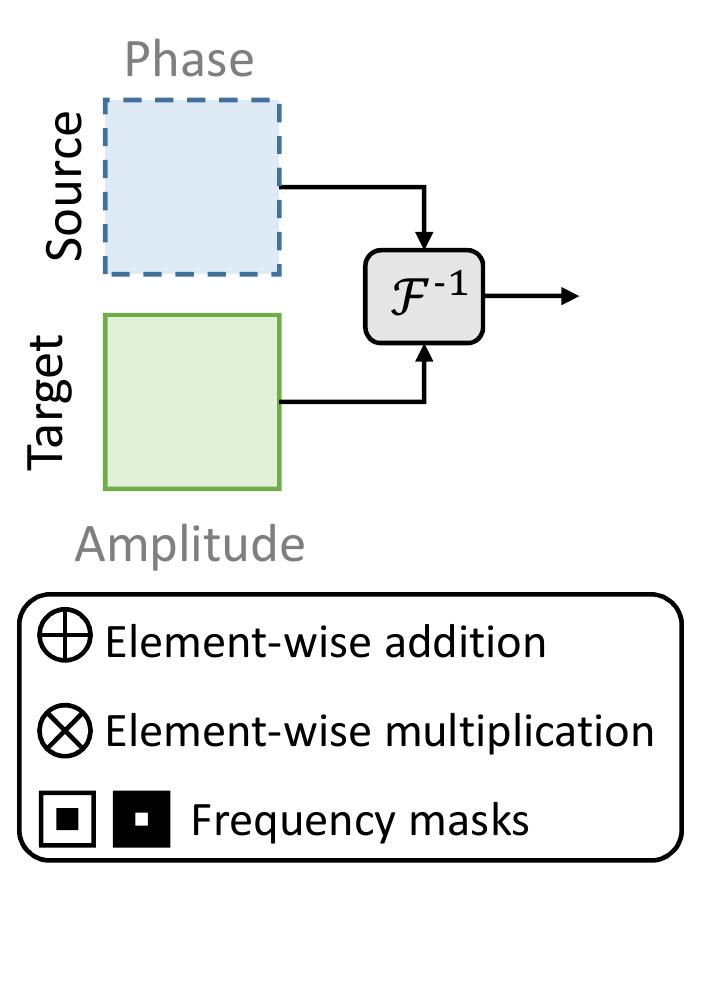}
        &\includegraphics[clip,trim= 0cm 8cm 19cm 2cm,width=0.247\textwidth]{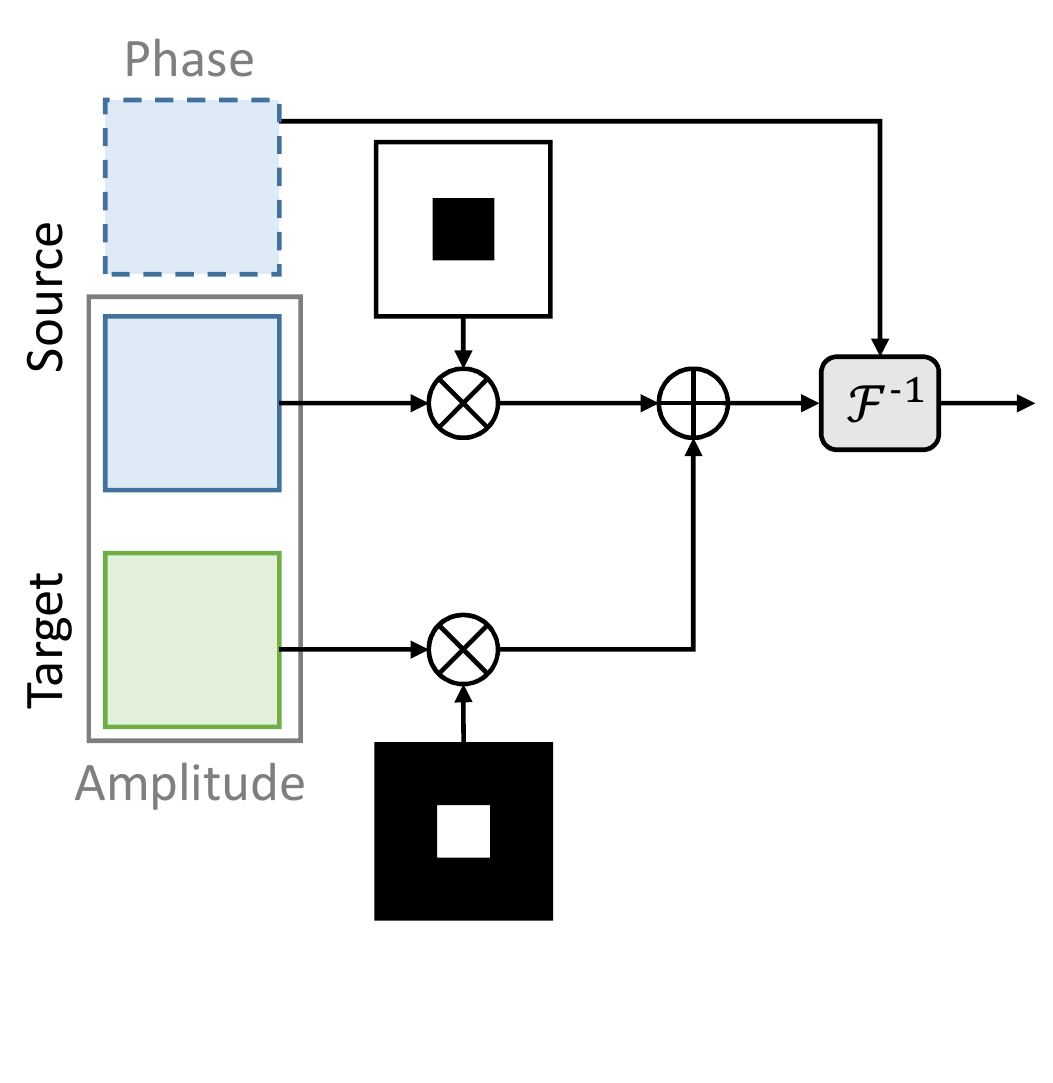}
        &\includegraphics[clip,trim= 0cm 8cm 21.5cm 2cm,width=0.2\textwidth]{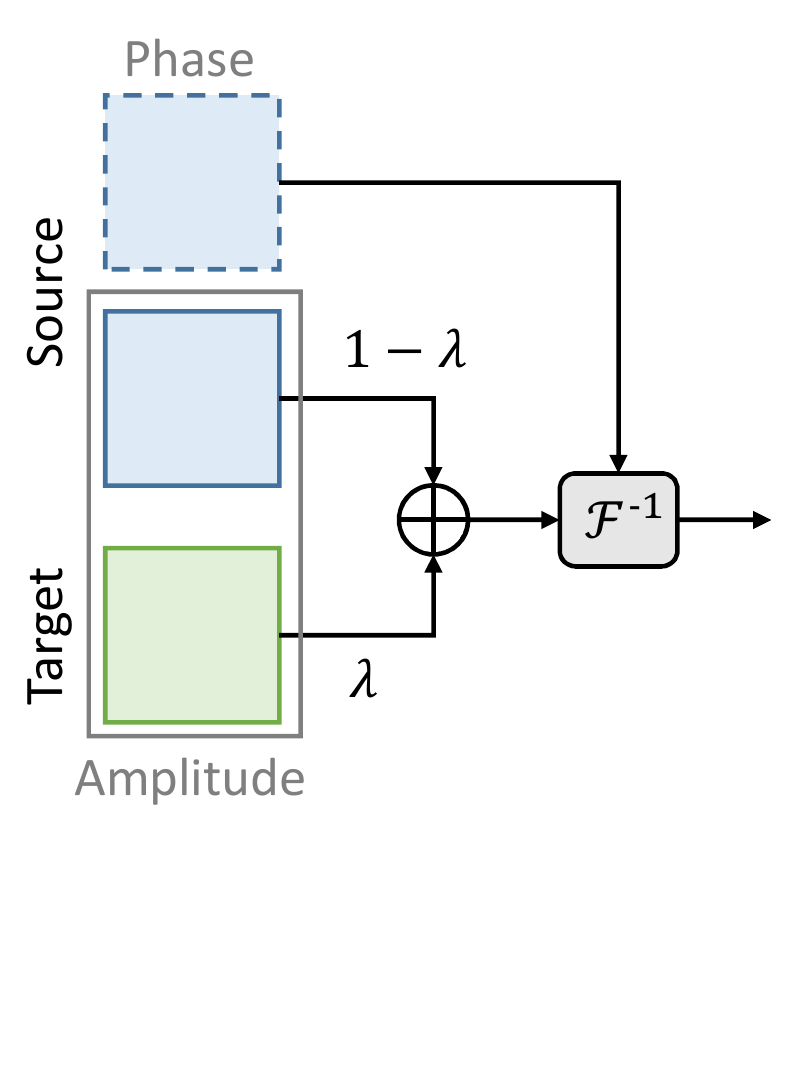}
        &\includegraphics[clip,trim= 0cm 8cm 17.5cm 2cm,width=0.3\textwidth]{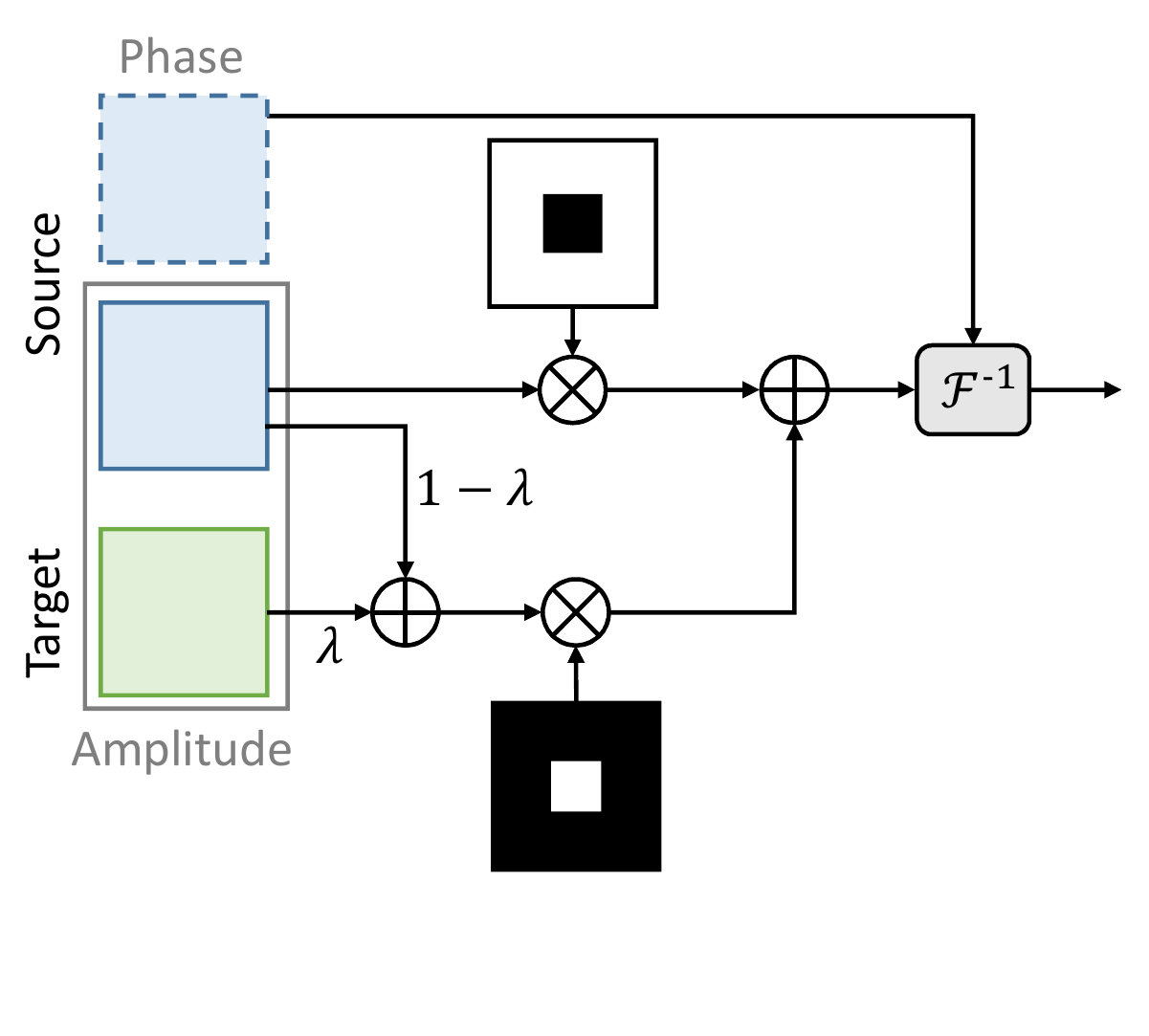}\\
        (a) Yang et al. \cite{yang2020phase} &(b) Yang et al. \cite{yang2020fda} &(c) Xu et al. \cite{xu2021fourier} &(d) Liu et al. \cite{liu2021feddg}\\
    \end{tabular}

\caption{Comparisons of different mixup strategies in frequency domain. Yang et al. \cite{yang2020phase} replaced the amplitude spectrum of the source image directly with the amplitude spectrum of the target image. Yang et al. \cite{yang2020fda} applied a frequency mask to swap the low-frequency components of the source amplitude spectrum. Xu et al. \cite{xu2021fourier} integrated two amplitude spectra by using a weighted sum operation. Liu et al. \cite{liu2021feddg} retained the high-frequency components of the source amplitude spectrum and combined the low-frequency components of the source amplitude spectrum with those of the target image.}
\label{fig:other_SMU}
\end{figure*}

The main objective of this study is to develop a privacy-preserving face recognition model by training it on a synthetic dataset.
To this end, we propose a novel data augmentation technique called Spectrum Mixup (SMU) that reduces the domain gap between real and synthetic datasets by immigrating information from the amplitude spectrum of real images.
In contrast to other mixup strategies in the frequency domain, as shown in Fig. \ref{fig:other_SMU}, that use weighted sum operation or hard-assignment mask, we integrate the amplitude components of synthetic data and the amplitude components of real data using a Gaussian-based soft-assignment map, and enhance high-frequency information, as illustrated in Fig. \ref{fig:SMU}.
Our approach is based on the following hypotheses: 1) semantic content (identity information) is mainly encoded in the phase components; 2) incorporating amplitude information from real data into synthetic data results in a better fit to the distribution of the real dataset; and 3) enhancing high-frequency information is more effective than low-frequency information since deep neural networks prioritize fitting certain frequencies \cite{xu2020frequency}, usually from low to high, which indicates that synthetic data carry realistic low-frequency information but lack high-frequency details.

To obtain the frequency components of an image\footnote{For simplicity, the single-channel image is used to illustrate the procedure of discrete Fourier transform, while the extension to color images is straightforward by processing on each channel separately using the same way.} $\textit{\textbf{x}} \in \mathbb{R}^{M \times N}$, we use the 2D discrete Fourier transform, which can be expressed as:

\begin{equation}
    \mathcal{F}(\textit{\textbf{x}})(u,v)=\sum^{M-1}_{m=0}\sum^{N-1}_{n=0}x(m,n)e^{-j2\pi(\frac{um}{M}+\frac{vn}{N})},
    \label{equ:DFT}
\end{equation}
where $(m,n)$ denotes the coordinate of an image pixel in the spatial domain; $x(m,n)$ is the pixel value; $(u,v)$ represents the coordinate of a spatial frequency in frequency domain; $\mathcal{F}(\textit{\textbf{x}})(u,v)$ is the complex frequency value of image \textit{\textbf{x}}; $e$ and $j$ are Euler's number and the imaginary unit, respectively. 
Accordingly, $\mathcal{F}^{-1}(\cdot)$ is the 2D inverse discrete Fourier transform which converts frequency spectrum to spatial domain.
Following Euler's formula:
\begin{equation}
    e^{j\theta}=\cos(\theta)+j\sin(\theta),
\end{equation}
the natural exponential function in Eq. (\ref{equ:DFT}) can be rewritten as:
\begin{equation}
    e^{-j2\pi\!(\frac{um}{M}\!+\!\frac{vn}{N})}\!=\!\cos2\pi\!\left(\frac{um}{M}\!+\!\frac{vn}{N}\right)\!-\!j\sin2\pi\!\left(\frac{um}{M}\!+\!\frac{vn}{N}\right).
    \label{equ:expo}
\end{equation}
According to Eq. (\ref{equ:DFT}) and Eq. (\ref{equ:expo}), the image is decomposed into orthogonal sine and cosine functions which constitute the imaginary and real part of the frequency component $\mathcal{F}(\textit{\textbf{x}})$, respectively.
Then, the amplitude and phase spectra of $\mathcal{F}(\textit{\textbf{x}})(u,v)$ are defined as:
\begin{equation}
    \mathcal{A}(\textit{\textbf{x}})(u,v)=\left(R^{2}(\textit{\textbf{x}})(u,v)+I^{2}(\textit{\textbf{x}})(u,v)\right)^{1/2},
\end{equation}
\begin{equation}
    \mathcal{P}(\textit{\textbf{x}})(u,v)=\arctan\left(\frac{I(\textit{\textbf{x}})(u,v)}{R(\textit{\textbf{x}})(u,v)}\right),
\end{equation}
where $R(\textit{\textbf{x}})$ and $I(\textit{\textbf{x}})$ represent the real part and imaginary part of $\mathcal{F}(\textit{\textbf{x}})$, respectively.

Furthermore, to implement our SMU method, we use a Gaussian kernel to create a soft-assignment map, denoted as $\textit{\textbf{G}}$. The soft-assignment map is defined as follows:
\begin{equation}
\textit{\textbf{G}}(u,v) = e^{-D^{2}(u,v)/2D^{2}_{0}},
\end{equation}
where $D_{0}$ is a positive constant that represents the cut-off frequency, and $D^{2}_{0}$ is the distance between a point $(u,v)$ in the frequency domain and the center of the frequency rectangle, that is,
\begin{equation}
D(u,v)=\left((u-M/2)^{2}+(v-N/2)^{2}\right)^{1/2},
\end{equation}
where $M$ and $N$ represent the height and width of the frequency rectangle and image, respectively.

The SMU procedure for two randomly sampled images $\textit{\textbf{x}}_{syn}$ and $\textit{\textbf{x}}_{real}$, can be formalized as follows:
\begin{equation}
    \textit{\textbf{x}}'_{syn}=\mathcal{F}^{-1}((\textbf{1}-\textit{\textbf{G}})\circ\mathcal{A}(\textit{\textbf{x}}_{real})+\textit{\textbf{G}}\circ\mathcal{A}(\textit{\textbf{x}}_{syn}),\mathcal{P}(\textit{\textbf{x}}_{syn})),
\end{equation}
where $\circ$ denotes the element-wise multiplication operation.
We maintain the low-frequency information of synthetic data and immigrate high-frequency details from the amplitude components of the real image.
The resulting amplitude components are then combined with the phase components of $\textit{\textbf{x}}_{syn}$ to obtain the final augmented synthetic image $\textit{\textbf{x}}'_{syn}$.

To summarize, the SMU procedure uses a soft-assignment map to combine the low-frequency components of the synthetic image with the high-frequency components of the real image, resulting in a more realistic augmented synthetic image.
Note that, the SMU method only uses the amplitude spectra of the real images to obtain high-frequency components, while the labels or identity information of the real images is not used during the training process.
It means that the method can be applied to any type of image dataset without the need for manual annotation or labeling, making it a useful tool for various applications in computer vision.

\section{Experiments}\label{sec:results}
\subsection{Experiment setup and evaluation protocol}
In this paper, LResNet50E-IR model \cite{deng2019arcface} is used as our backbone for face recognition.
We conduct all experiments using the learning strategies and experimental hyperparameters of the state-of-the-art methods \cite{bae2023digiface,qiu2021synface} on 4 NVIDIA V100 GPUs.
The batch size is 256, and the number of epochs is 40.
The initial learning rate is 0.1, and it is divided by 10 at 24-th, 30-th, and 36-th epoch.
The loss function is Adaface loss \cite{kim2022adaface}, and SGD optimizer is used to train models.
For a fair and steady comparison, our model is trained from scratch and no early stop strategy is used in our experiments.
The trained model at the last epoch is used to conduct inference on the testing datasets.

\subsection{Datasets}
In this study, the Digiface1M (synthetic) dataset \cite{bae2023digiface} and the CASIA-WebFace (real) dataset \cite{yi2014learning_casia_webface} are used to train our face recognition model.
For the Digiface1M dataset, there are 720k face images with a resolution of $256 \times 256$ in total, in which each identity consists of 72 images.
For the CASIA-WebFace dataset, it consists of 528k images with a resolution of $112 \times 112$ preprocessed by \cite{deng2019arcface} for 10575 identities.
Note that only face images from the CASIA-WebFace dataset are used in the proposed SMU module; therefore, the identity information is not adopted in our training stage.
For preprocessing, we randomly select 200 identities with 50 face images as our real image data on the CASIA-WebFace, and crop all of the images into a size of $112 \times 112$.

To evaluate the recognition performance, we test on five common face verification benchmarks, including LFW \cite{huang2008labeled}, AgeDB-30 \cite{moschoglou2017agedb}, CALFW \cite{zheng2017calfw}, CFP-FP \cite{sengupta2016cfp-fp}, and CP-LFW \cite{zheng2018cplfw} datasets, and compute verification accuracy for evaluation.
The LFW is one of the most common benchmark datasets which contains 6000 pairs of images collected in-the-wild.
The AgeDB-30 and CALFW aim to test the model performance under large age variation.
The CFP-FP and CP-LFW datasets are used to evaluate recognition ability with large pose variation.


\subsection{Choice of synthetic dataset}
Recently, a study \cite{zhu2021webface260m} reports that a higher number of identities and image samples can improve recognition performance with diverse and robust embedding.
Bae et al. \cite{bae2023digiface} demonstrate that the robustness and efficiency of their pipeline can generate face images with the same identity in any conditions, and reveal a high correlation between image number and performance.
Also, the experimental results of several studies \cite{qiu2021synface, deng2020discofacegan} suggest a similar conclusion to the above studies.
However, there are significant biases between face images when considering the consistency of intra-identity images.
These biases lead to difficulty recognizing whether they are the same identity even if generated by the same identity information, as shown in Fig.~\ref{fig:discofacegan_vs_digiface1m}.
We further conduct experiments to investigate the impact of sample numbers for each identity.
As shown in Table~\ref{tab:discofacegan impact of sample increase}, higher sample numbers with identity inconsistency affect the recognition performance, gradually.
In this way, we choose the Digiface1M as our training dataset to develop a face recognition system.


\begin{table}[t]
\centering
\small
\caption{The impact of sample number for each identity in DiscofaceGAN~\cite{deng2020discofacegan}. We have tested different settings with the default Adaface training processing~\cite{kim2022adaface} without any other data augmentation or modification methods.}
\begin{tabular}{@{}c@{ \ }c@{ \ }c@{ \ }c@{ \ }c@{ \ }c@{ \ }c@{}}
\hline
\#Sample & LFW   & AgeDB-30 & CA-LFW & CFP-FP & CP-LFW & Avg.   \\ \hline
40       & \textbf{90.68} & \textbf{65.67}    & \textbf{75.55}  & \textbf{68.59}  & 68.40  & \textbf{73.78} \\
50       & 88.90 & 65.45    & 74.82  & 68.57  & \textbf{68.63}  & 73.27 \\
72       & 89.50 & 63.55    & 73.63  & 67.04  & 67.50  & 72.25 \\
100      & 88.95 & 63.17    & 73.52  & 66.01  & 66.48  & 71.63 \\
200      & 88.32 & 61.63    & 71.17  & 65.43  & 66.05  & 70.52 \\ \hline
\end{tabular}
\label{tab:discofacegan impact of sample increase}
\end{table}

\subsection{Ablation studies}
\subsubsection{SA selection}\label{sec:abl_DA}
In this study, we conduct several experiments to find the best combination of data augmentations for performance improvement.
All details of the data augmentation setting are reported in our supplementary document.
Finally, our combination of data augmentations consists of low-resolution operation, random cropping operation, photometric augmentation, grayscale operation, and perspective operations.
As shown in Table \ref{tab:abl_data_augmentation} and Table \ref{table: additional data augmentation}, the experimental results suggest that better performance can be obtained when increasing the strength (probability) of data augmentation.

\begin{table*}[t]
    \centering
    \caption{Notations and settings for SA. $p^*$ denote the probability of adopting data augmentation.}
    \begin{tabular}{c|l|ccccccc}
    \toprule
    Name  & Description & $p^{LR}$  & $p^{Crop}$ & $p^{Pho}$ & $p^{Gray}$ & $p^{Per}$ & $p^{GB}$  & $p^{GN}$  \\ \midrule
    DA-S0 & Original in \cite{kim2022adaface}          & 0.2 & 0.2  & 0.2 & -      &  -           & -   &  -  \\
    DA-S1 & Weakest SA          & 0.2 & 0.2  & 0.2 & 0.01      &  -           & -   &  -  \\
    DA-S2 & Forth strongest SA  & 0.5 & 0.5  & 0.5 & 0.2       &  -           & -   &  -  \\
    DA-S3 & Third strongest SA  & 0.5 & 0.5  & 0.5 & 0.4       &  -          &  -  &  -  \\
    DA-S4 & Second strongest SA & 0.5 & 0.5  & 0.5 & 0.4       & 0.4         &  -  &  -  \\
    DA-S5 & Strongest SA        & 0.5 & 0.5  & 0.5 & 0.4       & 0.4         & 0.4 & 0.4 \\ \bottomrule
    \end{tabular}
    \label{table: additional data augmentation}
\end{table*}

\begin{table}[t]
\centering
\small
\caption{Ablation study of the SA method.} 
    \begin{tabular}{@{}l@{ \ }c@{ \ }c@{ \ }c@{ \ }c@{ \ }c@{ \ }c@{}}
    \hline
    Data Aug.                  & LFW & AgeDB-30 & CA-LFW & CFP-FP & CP-LFW & Avg. \\ \hline
    -                          & 88.43 & 67.27    & 71.52  & 76.57  &  69.45 & 74.65 \\ 
    {DA-S1}                    & 89.02 & 71.63    & 72.85  & 77.93  & 69.75  & 76.24 \\ 
    {DA-S2}                        & 91.55 & 76.72    & 77.12  & 81.27  & 73.05  & 79.94 \\ 
    {DA-S3}                        & 91.83 & 75.23    & 77.38  & 81.09  & 73.67  & 79.84 \\  
    {DA-S4}                        & 92.72 & 75.43    & 77.60  & 82.99  & 74.53  & 80.65 \\ 
    {DA-S5}                        & \textbf{94.00} & \textbf{77.75}    & \textbf{79.97}  & \textbf{84.17}  & \textbf{78.38}  & \textbf{82.85} \\ 
    \hline
    \end{tabular}
\label{tab:abl_data_augmentation}
\end{table}

\subsubsection{Impact of the cut-off frequency in SASMU}
In this experiment, the proposed SASMU method has been tested with different settings of the cut-off frequency $D_{0}$ for performance evaluation.
As shown in Table \ref{tab:abl_SMU}, the average accuracy is 83.96\% without the SMU operation, and the performance can be improved by using proposed method except for $D_{0}=15$.
The best average accuracy (84.56\%) is yielded by the proposed method with $D_{0}=60$.

\begin{table}[t]
\small
\caption{Ablation study of the proposed SASMU method for cut-off frequency $D_{0}$ of Gaussian kernel (\%). $S$ denotes a sample space ($S=\{15, 30, 45, 60\}$ is used in this experiment) and $U$ stands for uniform distribution. The best average accuracy is 84.56\% with $D_{0}=60$.} 
    \begin{tabular}{@{}l@{ \ }c@{ \ }c@{ \ }c@{ \ }c@{ \ }c@{ \ }c@{}}
    \hline
    Cut-off freq.                  & \footnotesize{LFW} & \footnotesize{AgeDB-30} & \footnotesize{CA-LFW} & \footnotesize{CFP-FP} & \footnotesize{CP-LFW} & \footnotesize{Avg.} \\ \hline
    \multicolumn{1}{l}{-}             & 95.32 & 78.37    & 81.37  & 85.21  & 79.53  & 83.96 \\ 
    $D_{0}=15$                        & 95.03 & 78.53    & 81.23  & 84.14  & 79.27  & 83.64 \\ 
    $D_{0}=30$                        & 95.72 & 79.75    & 81.72  & 85.33  & 79.78  & 84.46 \\  
    $D_{0}=45$                        & \textbf{95.77} & 78.90    & \textbf{82.32}  & 85.60  & \textbf{80.18}  & 84.55 \\ 
    $D_{0}=60$                        & 95.75 & \textbf{79.72}    & 81.97  & \textbf{85.63}  & 79.75  & \textbf{84.56} \\ 
    $D_{0}\sim U(S)$ & 95.72 & 79.32    & 81.57  & 84.97  & 80.33  & 84.38 \\ \hline
    \end{tabular}
\label{tab:abl_SMU}
\end{table}

\subsection{Quantitative results}
\subsubsection{Performance of frequency-based mixup methods}
In this experiment, we compare the proposed method to other mixup methods in frequency domain.
For fair comparison, we adopted SA method for all experiments.
As shown in Table \ref{tab:SOTA_SMU}, we obtain the best accuracy on all datasets when using our SASMU method.
It is noteworthy that the accuracy of other methods are lower than baseline which is without any mixup operation in frequency domain.
The reason might be that they assumed that semantic information is mainly encoded in high-frequency space, and thus preserved them while combining the low-frequency components of the target image for domain adaptation.
However, previous studies have claimed that there are serious domain gaps between real and synthetic data spaces, especially in frequency domain \cite{jiang2021focal, durall2020watch}, and learning high-frequency information is difficult than low-frequency \cite{xu2020frequency}.
This is the reason that we decide to immigrate the high-frequency information from the real space to synthetic data for minimizing the synthetic-to-real domain gap, instead of relying on the low-frequency components.

\begin{table}[t]
\centering
\small
\caption{Comparison of different mixup method in frequency domain (\%). The SA method is used in all experiments. Our method achieves the best performance on all datasets than other methods.}

\begin{tabular}{@{}l@{ \ }c@{ \ }c@{ \ }c@{ \ }c@{ \ }c@{ \ }c@{}}
    \hline
    Method                           & \footnotesize{LFW} & \footnotesize{AgeDB-30} & \footnotesize{CA-LFW} & \footnotesize{CFP-FP} & \footnotesize{CP-LFW} & \footnotesize{Avg.} \\ \hline
    -                                & 95.32 & 78.37 & 81.37 & 85.21 & 79.53 & 83.96 \\
    \footnotesize{Yang et al. \cite{yang2020phase}} & 92.68 & 74.93 & 78.58 & 80.44 & 73.82 & 80.09 \\
    \footnotesize{Yang et al. \cite{yang2020fda}}   & 94.10 & 77.97 & 80.30 & 82.67 & 75.08 & 82.02 \\ 
    \footnotesize{Xu et al. \cite{xu2021fourier}}   & 94.83 & 78.20 & 80.27 & 84.41 & 77.17 & 82.98 \\ 
    \footnotesize{Liu et al. \cite{liu2021feddg}}   & 94.93 & 77.90 & 80.17 & 83.79 & 77.75 & 82.91 \\ \hline
    Ours & \textbf{95.75} & \textbf{79.72} & \textbf{81.97} & \textbf{85.63} & \textbf{79.75}  & \textbf{84.56} \\
    \hline
\end{tabular}

\label{tab:SOTA_SMU}
\end{table}

\subsubsection{Visualization of frequency-based mixup methods}
\begin{figure*}[t]
\centering
    \begin{tabular}{l}
        \includegraphics[clip,trim= 8cm 3.5cm 6.5cm 2.5cm,width=\linewidth]{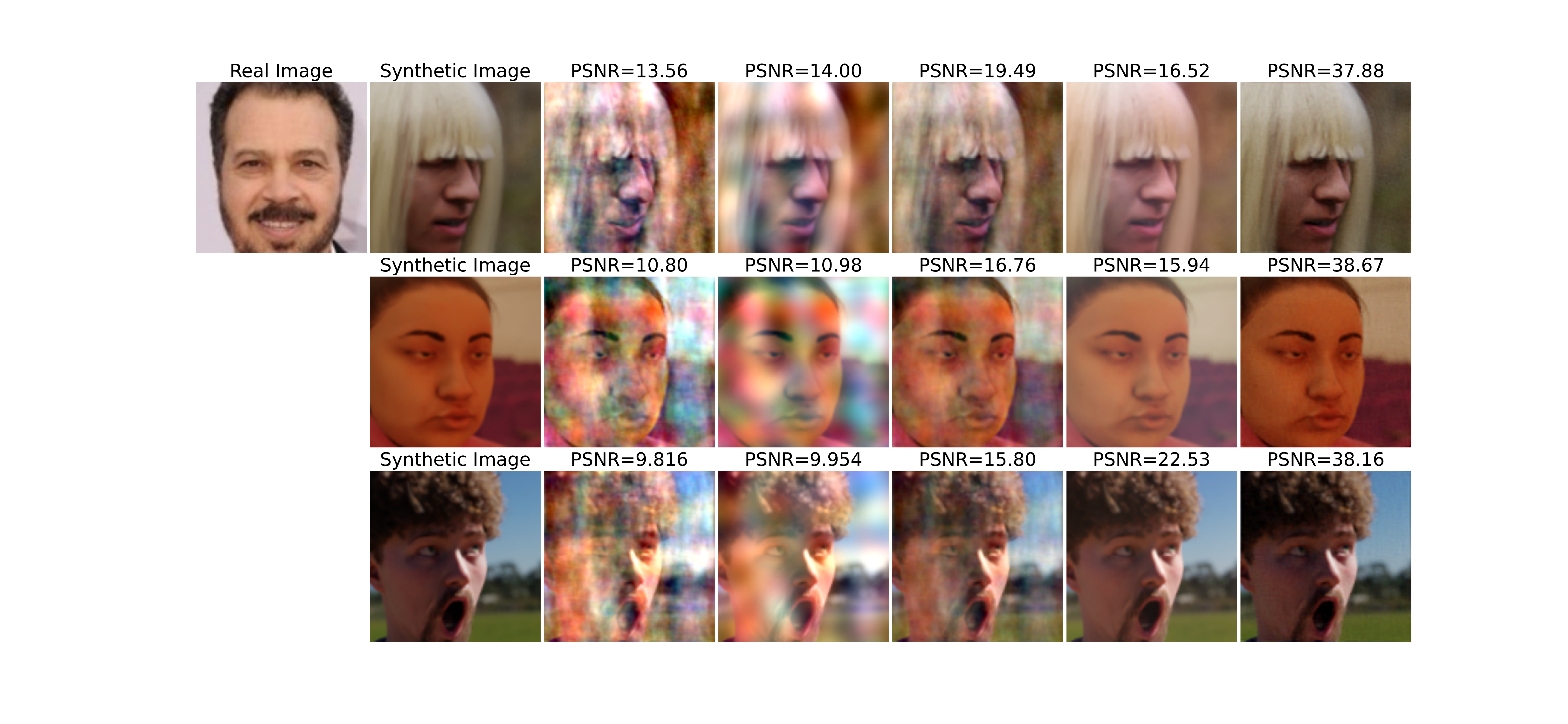}\\
       \qquad\qquad\qquad\qquad\qquad\qquad\qquad\ \ \ Yang et al. \cite{yang2020phase} \ \ Yang et al. \cite{yang2020fda} \quad Xu et al. \cite{xu2021fourier} \quad\ \! Liu et al. \cite{liu2021feddg} \qquad\quad\ Ours \\
    \end{tabular}
\caption{The visualization results using different mixup strategies in frequency domain and PSNR values indicating the image quality and similarity between original synthetic image and the augmented image.}
\label{fig:vis_comparison}
\end{figure*}

To investigate the effect of mixup operations in frequency domain, we have visualized the augmented images using different methods, as shown in Fig. \ref{fig:vis_comparison}.
The amplitude spectrum of synthetic images are combined with the amplitude spectrum of real image using the proposed SMU method and other methods.
In these experiments, the optimal hyperparameters of other mixup methods are used.
Yang et al. \cite{yang2020phase} directly replace the amplitude of the synthetic image with that of the real image, which causes the inconsistency between the phase and amplitude of the synthetic image.
Yang et al. \cite{yang2020fda} swap low-frequency components using a square mask between two images, leading to a ringing effect on the augmented image as the square mask works as an ideal filter.
Xu et al. \cite{xu2021fourier} adopt weighted sum operation to combine amplitude spectra, without considering that different frequencies have different importance and information, which produces artifacts in the augmented images.
Liu et al. \cite{liu2021feddg} retain the high-frequency components of the synthetic amplitude spectrum and combined the low-frequency components of the source amplitude spectrum with those of the real image.
However, their setting leads to only a few frequency points being adjusted on the synthetic image, resulting in only image intensities being changed in the spatial domain, merely.
In other words, when enlarging the hyperparameter of their method, it will cause the ringing effect which is in line with the results of \cite{yang2020fda}.
In addition, we compute the peak signal-to-noise ratio (PSNR) values of those augmented images and show them in Fig. \ref{fig:vis_comparison}. 
It indicates that our method can produce high-quality images which are similar to the original synthetic images.

\section{Conclusion}\label{sec:conclusions}
In this study, we have investigated how to develop a robust face recognition system with consideration for the privacy-preserving issue using the synthetic dataset.
The proposed SASMU method is used to increase data variation and minimize the synthetic-to-real domain gap, which consists of the best combination of spatial data augmentations (SA) and spectrum mixup (SMU).
First, we have analyzed how common data augmentations improve the recognition model for considering different conditions in the real scene and various color spaces (e.g., RGB-/gray-space), and have found the best combination of data augmentations for face recognition when using synthetic dataset.
Second, we have investigated the reason for causing the  domain gap between real and synthetic datasets, and have proposed a novel mixup method on the frequency domain, SMU, to reduce the gap for improving recognition performance.
Note that only synthetic data and real images (without labels) are used, and no data from the target dataset is used in the training stage.
Extensive experimental results have demonstrated the effectiveness of SASMU, achieving state-of-the-art performance on several common face verification benchmarks, including LFW, AgeDB-30, CA-LFW, CFP-FP, and CP-LFW.
Our experimental results suggest that 1) SASMU is a crucial and efficient training strategy for face recognition; 2) applying SA can improve the recognition performance, especially using synthetic data; and 3) using the SMU method to immigrate high-frequency information from real data outperforms other methods with the opposite assumptions.
For future work, we can further discuss and analyze heavy data augmentation for face recognition or other tasks.
Furthermore, the SMU method may be applied to other computer vision tasks for domain generalization and may be used to develop a generalized foundation model for several studies.


{\small
\bibliographystyle{ieee_fullname}
\bibliography{egbib}

\begin{thebibliography}{10}\itemsep=-1pt

\bibitem{an2022killing}
Xiang An, Jiankang Deng, Jia Guo, Ziyong Feng, XuHan Zhu, Jing Yang, and
  Tongliang Liu.
\newblock Killing two birds with one stone: Efficient and robust training of
  face recognition cnns by partial fc.
\newblock In {\em Proceedings of the IEEE/CVF Conference on Computer Vision and
  Pattern Recognition}, pages 4042--4051, 2022.

\bibitem{bae2023digiface}
Gwangbin Bae, Martin de La~Gorce, Tadas Baltru{\v{s}}aitis, Charlie Hewitt,
  Dong Chen, Julien Valentin, Roberto Cipolla, and Jingjing Shen.
\newblock Digiface-1m: 1 million digital face images for face recognition.
\newblock In {\em Proceedings of the IEEE/CVF Winter Conference on Applications
  of Computer Vision}, pages 3526--3535, 2023.

\bibitem{boutros2023USynthFace}
Fadi Boutros, Marcel Klemt, Meiling Fang, Arjan Kuijper, and Naser Damer.
\newblock Unsupervised face recognition using unlabeled synthetic data.
\newblock In {\em 2023 IEEE 17th International Conference on Automatic Face and
  Gesture Recognition (FG)}, pages 1--8. IEEE, 2023.

\bibitem{chamikara2020ppfr_diffential_privacy}
Mahawaga Arachchige~Pathum Chamikara, Peter Bertok, Ibrahim Khalil, Dongxi Liu,
  and Seyit Camtepe.
\newblock Privacy preserving face recognition utilizing differential privacy.
\newblock {\em Computers \& Security}, 97:101951, 2020.

\bibitem{deng2021masked}
Jiankang Deng, Jia Guo, Xiang An, Zheng Zhu, and Stefanos Zafeiriou.
\newblock Masked face recognition challenge: The insightface track report.
\newblock In {\em Proceedings of the IEEE/CVF International Conference on
  Computer Vision}, pages 1437--1444, 2021.

\bibitem{deng2019arcface}
Jiankang Deng, Jia Guo, Niannan Xue, and Stefanos Zafeiriou.
\newblock Arcface: Additive angular margin loss for deep face recognition.
\newblock In {\em Proceedings of the IEEE/CVF conference on computer vision and
  pattern recognition}, pages 4690--4699, 2019.

\bibitem{deng2021variational}
Jiankang Deng, Jia Guo, Jing Yang, Alexandros Lattas, and Stefanos Zafeiriou.
\newblock Variational prototype learning for deep face recognition.
\newblock In {\em Proceedings of the IEEE/CVF Conference on Computer Vision and
  Pattern Recognition}, pages 11906--11915, 2021.

\bibitem{deng2020discofacegan}
Yu Deng, Jiaolong Yang, Dong Chen, Fang Wen, and Xin Tong.
\newblock Disentangled and controllable face image generation via 3d
  imitative-contrastive learning.
\newblock In {\em Proceedings of the IEEE/CVF conference on computer vision and
  pattern recognition}, pages 5154--5163, 2020.

\bibitem{durall2020watch}
Ricard Durall, Margret Keuper, and Janis Keuper.
\newblock Watch your up-convolution: Cnn based generative deep neural networks
  are failing to reproduce spectral distributions.
\newblock In {\em Proceedings of the IEEE/CVF conference on computer vision and
  pattern recognition}, pages 7890--7899, 2020.

\bibitem{guo2016ms}
Yandong Guo, Lei Zhang, Yuxiao Hu, Xiaodong He, and Jianfeng Gao.
\newblock Ms-celeb-1m: A dataset and benchmark for large-scale face
  recognition.
\newblock In {\em European conference on computer vision}, pages 87--102.
  Springer, 2016.

\bibitem{huang2008labeled}
Gary~B Huang, Marwan Mattar, Tamara Berg, and Eric Learned-Miller.
\newblock Labeled faces in the wild: A database forstudying face recognition in
  unconstrained environments.
\newblock In {\em Workshop on faces in'Real-Life'Images: detection, alignment,
  and recognition}, 2008.

\bibitem{huang2021age}
Zhizhong Huang, Junping Zhang, and Hongming Shan.
\newblock When age-invariant face recognition meets face age synthesis: A
  multi-task learning framework.
\newblock In {\em Proceedings of the IEEE/CVF conference on computer vision and
  pattern recognition}, pages 7282--7291, 2021.

\bibitem{ji2022ppfr_learned_budgets}
Jiazhen Ji, Huan Wang, Yuge Huang, Jiaxiang Wu, Xingkun Xu, Shouhong Ding,
  ShengChuan Zhang, Liujuan Cao, and Rongrong Ji.
\newblock Privacy-preserving face recognition with learnable privacy budgets in
  frequency domain.
\newblock In {\em Computer Vision--ECCV 2022: 17th European Conference, Tel
  Aviv, Israel, October 23--27, 2022, Proceedings, Part XII}, pages 475--491.
  Springer, 2022.

\bibitem{jiang2021focal}
Liming Jiang, Bo Dai, Wayne Wu, and Chen~Change Loy.
\newblock Focal frequency loss for image reconstruction and synthesis.
\newblock In {\em Proceedings of the IEEE/CVF International Conference on
  Computer Vision}, pages 13919--13929, 2021.

\bibitem{karkkainen2021fairface}
Kimmo Karkkainen and Jungseock Joo.
\newblock Fairface: Face attribute dataset for balanced race, gender, and age
  for bias measurement and mitigation.
\newblock In {\em Proceedings of the IEEE/CVF Winter Conference on Applications
  of Computer Vision}, pages 1548--1558, 2021.

\bibitem{kim2019fine_tune_nir}
Jeyeon Kim, Hoon Jo, Moonsoo Ra, and Whoi-Yul Kim.
\newblock Fine-tuning approach to nir face recognition.
\newblock In {\em ICASSP 2019-2019 IEEE International Conference on Acoustics,
  Speech and Signal Processing (ICASSP)}, pages 2337--2341. IEEE, 2019.

\bibitem{kim2022adaface}
Minchul Kim, Anil~K Jain, and Xiaoming Liu.
\newblock Adaface: Quality adaptive margin for face recognition.
\newblock In {\em Proceedings of the IEEE/CVF Conference on Computer Vision and
  Pattern Recognition}, pages 18750--18759, 2022.

\bibitem{li2021spherical}
Shen Li, Jianqing Xu, Xiaqing Xu, Pengcheng Shen, Shaoxin Li, and Bryan Hooi.
\newblock Spherical confidence learning for face recognition.
\newblock In {\em Proceedings of the IEEE/CVF Conference on Computer Vision and
  Pattern Recognition}, pages 15629--15637, 2021.

\bibitem{li2018occlusion_CNN_attention}
Yong Li, Jiabei Zeng, Shiguang Shan, and Xilin Chen.
\newblock Occlusion aware facial expression recognition using cnn with
  attention mechanism.
\newblock {\em IEEE Transactions on Image Processing}, 28(5):2439--2450, 2018.

\bibitem{lin2021xcos}
Yu-Sheng Lin, Zhe-Yu Liu, Yu-An Chen, Yu-Siang Wang, Ya-Liang Chang, and
  Winston~H Hsu.
\newblock xcos: An explainable cosine metric for face verification task.
\newblock {\em ACM Transactions on Multimedia Computing, Communications, and
  Applications (TOMM)}, 17(3s):1--16, 2021.

\bibitem{liu2021feddg}
Quande Liu, Cheng Chen, Jing Qin, Qi Dou, and Pheng-Ann Heng.
\newblock Feddg: Federated domain generalization on medical image segmentation
  via episodic learning in continuous frequency space.
\newblock In {\em Proceedings of the IEEE/CVF Conference on Computer Vision and
  Pattern Recognition}, pages 1013--1023, 2021.

\bibitem{Liu_2017_sphereface}
Weiyang Liu, Yandong Wen, Zhiding Yu, Ming Li, Bhiksha Raj, and Le Song.
\newblock Sphereface: Deep hypersphere embedding for face recognition.
\newblock In {\em The IEEE Conference on Computer Vision and Pattern
  Recognition (CVPR)}, 2017.

\bibitem{maze2018iarpa}
Brianna Maze, Jocelyn Adams, James~A Duncan, Nathan Kalka, Tim Miller, Charles
  Otto, Anil~K Jain, W~Tyler Niggel, Janet Anderson, Jordan Cheney, et~al.
\newblock Iarpa janus benchmark-c: Face dataset and protocol.
\newblock In {\em 2018 international conference on biometrics (ICB)}, pages
  158--165. IEEE, 2018.

\bibitem{meng2021magface}
Qiang Meng, Shichao Zhao, Zhida Huang, and Feng Zhou.
\newblock Magface: A universal representation for face recognition and quality
  assessment.
\newblock In {\em Proceedings of the IEEE/CVF Conference on Computer Vision and
  Pattern Recognition}, pages 14225--14234, 2021.

\bibitem{miyamoto2021joint_nir_vis}
Takaya Miyamoto, Hiroshi Hashimoto, Akihiro Hayasaka, Akinori~F Ebihara, and
  Hitoshi Imaoka.
\newblock Joint feature distribution alignment learning for nir-vis and vis-vis
  face recognition.
\newblock In {\em 2021 IEEE International Joint Conference on Biometrics
  (IJCB)}, pages 1--8. IEEE, 2021.

\bibitem{moschoglou2017agedb}
Stylianos Moschoglou, Athanasios Papaioannou, Christos Sagonas, Jiankang Deng,
  Irene Kotsia, and Stefanos Zafeiriou.
\newblock Agedb: the first manually collected, in-the-wild age database.
\newblock In {\em proceedings of the IEEE conference on computer vision and
  pattern recognition workshops}, pages 51--59, 2017.

\bibitem{neto2021focusface}
Pedro~C Neto, Fadi Boutros, Jo{\~a}o~Ribeiro Pinto, Naser Damer, Ana~F
  Sequeira, and Jaime~S Cardoso.
\newblock Focusface: Multi-task contrastive learning for masked face
  recognition.
\newblock In {\em 2021 16th IEEE International Conference on Automatic Face and
  Gesture Recognition (FG 2021)}, pages 01--08. IEEE, 2021.

\bibitem{qiu2021synface}
Haibo Qiu, Baosheng Yu, Dihong Gong, Zhifeng Li, Wei Liu, and Dacheng Tao.
\newblock Synface: Face recognition with synthetic data.
\newblock In {\em Proceedings of the IEEE/CVF International Conference on
  Computer Vision}, pages 10880--10890, 2021.

\bibitem{sengupta2016cfp-fp}
Soumyadip Sengupta, Jun-Cheng Chen, Carlos Castillo, Vishal~M Patel, Rama
  Chellappa, and David~W Jacobs.
\newblock Frontal to profile face verification in the wild.
\newblock In {\em 2016 IEEE winter conference on applications of computer
  vision (WACV)}, pages 1--9. IEEE, 2016.

\bibitem{shorten2019data_augmentation_survey}
Connor Shorten and Taghi~M Khoshgoftaar.
\newblock A survey on image data augmentation for deep learning.
\newblock {\em Journal of big data}, 6(1):1--48, 2019.

\bibitem{sun2020circle}
Yifan Sun, Changmao Cheng, Yuhan Zhang, Chi Zhang, Liang Zheng, Zhongdao Wang,
  and Yichen Wei.
\newblock Circle loss: A unified perspective of pair similarity optimization.
\newblock In {\em Proceedings of the IEEE/CVF Conference on Computer Vision and
  Pattern Recognition}, pages 6398--6407, 2020.

\bibitem{terhorst2023qmagface}
Philipp Terh{\"o}rst, Malte Ihlefeld, Marco Huber, Naser Damer, Florian
  Kirchbuchner, Kiran Raja, and Arjan Kuijper.
\newblock Qmagface: Simple and accurate quality-aware face recognition.
\newblock In {\em Proceedings of the IEEE/CVF Winter Conference on Applications
  of Computer Vision}, pages 3484--3494, 2023.

\bibitem{wang2017normface}
Feng Wang, Xiang Xiang, Jian Cheng, and Alan~Loddon Yuille.
\newblock Normface: L2 hypersphere embedding for face verification.
\newblock In {\em Proceedings of the 25th ACM international conference on
  Multimedia}, pages 1041--1049, 2017.

\bibitem{wang2018cosface}
Hao Wang, Yitong Wang, Zheng Zhou, Xing Ji, Dihong Gong, Jingchao Zhou, Zhifeng
  Li, and Wei Liu.
\newblock Cosface: Large margin cosine loss for deep face recognition.
\newblock In {\em Proceedings of the IEEE conference on computer vision and
  pattern recognition}, pages 5265--5274, 2018.

\bibitem{wang2020facial_cyclegan}
Huijiao Wang, Haijian Zhang, Lei Yu, Li Wang, and Xulei Yang.
\newblock Facial feature embedded cyclegan for vis-nir translation.
\newblock In {\em ICASSP 2020-2020 IEEE International Conference on Acoustics,
  Speech and Signal Processing (ICASSP)}, pages 1903--1907. IEEE, 2020.

\bibitem{wang2022facemae}
Kai Wang, Bo Zhao, Xiangyu Peng, Zheng Zhu, Jiankang Deng, Xinchao Wang, Hakan
  Bilen, and Yang You.
\newblock Facemae: Privacy-preserving face recognition via masked autoencoders.
\newblock {\em arXiv preprint arXiv:2205.11090}, 2022.

\bibitem{wang2019rfw}
Mei Wang, Weihong Deng, Jiani Hu, Xunqiang Tao, and Yaohai Huang.
\newblock Racial faces in the wild: Reducing racial bias by information
  maximization adaptation network.
\newblock In {\em Proceedings of the ieee/cvf international conference on
  computer vision}, pages 692--702, 2019.

\bibitem{wang2021aan_face}
Qiangchang Wang and Guodong Guo.
\newblock Aan-face: attention augmented networks for face recognition.
\newblock {\em IEEE Transactions on Image Processing}, 30:7636--7648, 2021.

\bibitem{wang2021dsa}
Qiangchang Wang and Guodong Guo.
\newblock Dsa-face: Diverse and sparse attentions for face recognition robust
  to pose variation and occlusion.
\newblock {\em IEEE Transactions on Information Forensics and Security},
  16:4534--4543, 2021.

\bibitem{wang2022cqa_face}
Qiangchang Wang and Guodong Guo.
\newblock Cqa-face: Contrastive quality-aware attentions for face recognition.
\newblock In {\em Proceedings of the AAAI Conference on Artificial
  Intelligence}, volume~36, pages 2504--2512, 2022.

\bibitem{wang2020hierarchical_attention}
Qiangchang Wang, Tianyi Wu, He Zheng, and Guodong Guo.
\newblock Hierarchical pyramid diverse attention networks for face recognition.
\newblock In {\em Proceedings of the IEEE/CVF conference on computer vision and
  pattern recognition}, pages 8326--8335, 2020.

\bibitem{xu2021fourier}
Qinwei Xu, Ruipeng Zhang, Ya Zhang, Yanfeng Wang, and Qi Tian.
\newblock A fourier-based framework for domain generalization.
\newblock In {\em Proceedings of the IEEE/CVF Conference on Computer Vision and
  Pattern Recognition}, pages 14383--14392, 2021.

\bibitem{xu2020frequency}
Zhi-Qin~John Xu, Yaoyu Zhang, Tao Luo, Yanyang Xiao, and Zheng Ma.
\newblock Frequency principle: Fourier analysis sheds light on deep neural
  networks.
\newblock {\em Communications in Computational Physics}, 28(5):1746--1767,
  2020.

\bibitem{yang2020phase}
Yanchao Yang, Dong Lao, Ganesh Sundaramoorthi, and Stefano Soatto.
\newblock Phase consistent ecological domain adaptation.
\newblock In {\em Proceedings of the IEEE/CVF Conference on Computer Vision and
  Pattern Recognition}, pages 9011--9020, 2020.

\bibitem{yang2020fda}
Yanchao Yang and Stefano Soatto.
\newblock Fda: Fourier domain adaptation for semantic segmentation.
\newblock In {\em Proceedings of the IEEE/CVF Conference on Computer Vision and
  Pattern Recognition}, pages 4085--4095, 2020.

\bibitem{yi2014learning_casia_webface}
Dong Yi, Zhen Lei, Shengcai Liao, and Stan~Z Li.
\newblock Learning face representation from scratch.
\newblock {\em arXiv preprint arXiv:1411.7923}, 2014.

\bibitem{yin2017multi}
Xi Yin and Xiaoming Liu.
\newblock Multi-task convolutional neural network for pose-invariant face
  recognition.
\newblock {\em IEEE Transactions on Image Processing}, 27(2):964--975, 2017.

\bibitem{zheng2018cplfw}
Tianyue Zheng and Weihong Deng.
\newblock Cross-pose lfw: A database for studying cross-pose face recognition
  in unconstrained environments.
\newblock {\em Beijing University of Posts and Telecommunications, Tech. Rep},
  5(7), 2018.

\bibitem{zheng2017calfw}
Tianyue Zheng, Weihong Deng, and Jiani Hu.
\newblock Cross-age lfw: A database for studying cross-age face recognition in
  unconstrained environments.
\newblock {\em arXiv preprint arXiv:1708.08197}, 2017.

\bibitem{zhu2021masked}
Zheng Zhu, Guan Huang, Jiankang Deng, Yun Ye, Junjie Huang, Xinze Chen, Jiagang
  Zhu, Tian Yang, Jia Guo, Jiwen Lu, et~al.
\newblock Masked face recognition challenge: The webface260m track report.
\newblock {\em arXiv preprint arXiv:2108.07189}, 2021.

\bibitem{zhu2021webface260m}
Zheng Zhu, Guan Huang, Jiankang Deng, Yun Ye, Junjie Huang, Xinze Chen, Jiagang
  Zhu, Tian Yang, Jiwen Lu, Dalong Du, et~al.
\newblock Webface260m: A benchmark unveiling the power of million-scale deep
  face recognition.
\newblock In {\em Proceedings of the IEEE/CVF Conference on Computer Vision and
  Pattern Recognition}, pages 10492--10502, 2021.

\end{thebibliography}
}



\section*{Supplementary material}

# Supplementary Material for
# SASMU: boost the performance of generalized recognition model using synthetic face dataset

Chia-Chun Chung[1*]  Pei-Chun Chang[2*]  Yong-Sheng Chen[2]  HaoYuan He[1]  Chinson Yeh[1†]

[1]oToBrite Electronics, Inc., Taiwan

[2]Department of Computer Science, National Yang Ming Chiao Tung University, Taiwan

{zivzhong, michaelhe, charlesyeh}@otobrite.com, {pcchang.cs05, yschen}@nycu.edu.tw

## 1. Comparison with SOTA

We not only discussed Digiface1M [1] but also compared our approach with state-of-the-art research on the DiscofaceGAN dataset [3]. The comparison results are presented in Table 1. To ensure a fair comparison with USynthFace [2], we trained our model using synthetic data with the same number of training images (400K) from the DiscofaceGAN dataset. It is worth noting that when considering the same training strategies, such as the number of epochs, USynthFace achieved an average accuracy of 77.19%, while reporting an accuracy of 78.30% in their study. Our method achieved competitive performance with an average accuracy of 78.19% compared to other methods.

## 2. Implementation details

### 2.1. SA implementation

There are various types of data augmentation in visual-related task [10, 7]. In order to maintain focus and facilitate implementation, our selection of candidate methods and probabilities is restricted to the following items as Table. 2.

### 2.2. SMU implementation

The PyTorch-style pseudocode for the proposed SMU is shown in Algorithm 1. SMU performs a mixup processing in frequency domain to enhance the reality of the synthetic image. In this algorithm, the inputs are synthetic image, real image and cut-off frequency. First, we compute a Gaussian-based soft-assignment map using function GLPF. Then, synthetic and real images are converted into frequency domain to obtain spectra using fast Fourier transform (FFT) [8]. The augmented amplitude spectrum is calculated by combining synthetic amplitude spectrum and real amplitude spectrum using a soft-assignment map. Finally, the augmented image is obtained combination of the augmented amplitude and synthetic phase spectra by inverse FFT.

## 3. SA experimental results

This section discusses the optimal choice for spatial data augmentation (SA), including which approaches to use and how to set their probabilities. In addition, we address the order in which to apply data augmentation, specifically considering whether grayscale and histogram equalization should be applied before or after other approaches.

### 3.1. Appearance methods

Following the data augmentation setting of Adaface [4], we combine **low resolution** (LR), **cropping** (crop), and **photometric** (Pho) as base augmentation combination (**SA-B0**) with the same probability of 0.2 in Table. 3. We also introduce a stronger base augmentation with probability of 0.5, which we named **SA-B1**. All of our data augmentation experiments base on these two base augmentations.

#### 3.1.1 Using grayscale or not

To begin, we investigate the impact of applying grayscale during our training stage. We have tested various combinations of settings, as shown in Table. 3 and Table. 11. Our experimental results demonstrate the significant benefits of using the grayscale method with both SA-B0 and SA-B1. These results suggest that it is advisable to include grayscale as a standard part of our SA pipeline.

#### 3.1.2 Using channel shuffle or not

Building on our previous findings, we further investigate the impact of channel shuffle as a data augmentation method. In addition to **SA-B0** and **SA-B1**, we also test **SA-B1-G1** as a new base augmentation, which we denote as **SA-B2**. The

---
*Equal contribution
†Corresponding author

Table 1: Comparison with state-of-the-art methods on DiscofaceGAN dataset.

| Method | #Images | LFW | AgeDB-30 | CA-LFW | CFP-FP | CP-LFW | Avg |
|---|---|---|---|---|---|---|---|
| SynFace [9] | 500K | 88.98 | - | - | - | - | - |
| SynFace (w/IM) [9] | 500K | 91.97 | - | - | - | - | - |
| USynthFace [2] | 400K | 92.23 | 71.62 | 78.56 | 77.05 | 72.03 | 78.30 |
| USynthFace (40-th epoch)[2] | 400K | 91.42 | 69.73 | 75.43 | 77.94 | 71.07 | 77.19 |
| Ours (SASMU) | 400K | **92.60** | **73.53** | **78.17** | 74.36 | **72.28** | 78.19 |

Table 2: The details of data augmentation. $p^*$ denote the probability of adopting data augmentation.

| | Method | Source and hyperparameter | Notation of probability |
|---|---|---|---|
| Appearance | Crop | Default in Adaface [4] | $p^{Crop}$ |
| | Low Resolution | Default in Adaface [4] | $p^{LR}$ |
| | Photometric | Default in Adaface [4] | $p^{Pho}$ |
| | Grayscale | Torchvision [6] built-in, remains 3 channels | $p^{Gray}$ |
| | Gaussian Noise | skimage [11] built-in, random_noise(mode="gaussian") | $p^{GN}$ |
| | Gaussian Blur | Torchvision [6] built-in, kernel size = (7, 7) | $p^{GB}$ |
| | Channel Shuffle | Randomize the order of RGB channels | $p^{CS}$ |
| | Histogram Equalization | Torchvision [6] built-in | $p^{HE}$ |
| | Autocontrast | Torchvision [6] built-in | $p^{AC}$ |
| Geometry | Random Affine | Torchvision [6] built-in, degrees=(-30,30), translate=(0, 0.5), scale=(0.4, 0.5), shear=(0,0) | $p^{RA}$ |
| | Random Perspective | Torchvision [6] built-in, distortion_scale=np.random.uniform(0.0, 0.5) | $p^{RP}$ |

combination settings for channel shuffle are recorded in Table 4, and the results are presented in Table 12. Although channel shuffle showed promising results when applied to **SA-B0** and **SA-B1**, it had a negative impact on the performance of **SA-B1-G1**. Given the importance of maintaining a stable and reliable data augmentation pipeline, we decided to remove channel shuffle from our list of SA methods.

### 3.1.3 Gaussian Noise and Gaussian Blur

Based on the results from section 3.1.1, we expand our SA pipeline to include Gaussian blur and Gaussian noise. We conduct the following experiment upon **SA-B2**. We have tested various combinations of Gaussian blur and Gaussian noise, and the experiment settings are presented in Table 5. The results are shown in Table 13. Although Gaussian blur had a more significant impact when applied individually than Gaussian noise, using these two techniques jointly in the **SA-B1-G1** pipeline still showed a considerable improvement over using Gaussian blur alone. Thus, we decided to pair Gaussian blur and Gaussian noise and add them to the SA pipeline with equal probability. In addition, we get better performance when using **SA-B1-G2** with $p^{GN} = 0.4$ and $p^{GB} = 0.4$ (**SA-B3-GB1-GN1**).

### 3.2. Geometry methods

Real-world images present not only appearance challenges but also distortion challenges. To enhance the robustness of our model when facing real-world images, we incorporated mimic distortion as a data augmentation method. To prevent over-distortion, we select either random perspective or random affine based on which method provides the greatest performance boost.

We select **SA-B1** from Section 3.1.1 and **SA-B3-GB1-GN1** from Section 3.1.3 as the new base augmentations, denoted as **SA-B4**. Table 6 shows the experiment settings, and

**Algorithm 1:** PyTorch-style pseudocode for SMU

```
def GLPF(H, W, D0):
    # H, W: image size
    # D0:  cut-off frequency

    for i in range(H):
        for j in range(W):
            G[i,j] = exp(-sqrt((i-H)**2+(j-W)**2)/(2*(D0**2)))
    return G

def SMU(syn_img, real_img, D0):
    # syn_img/real_img:  synthetic/real image
    # D0:  cut-off frequency

    H, W = size(syn_img)
    G = GLPF(H, W, D0)

    # convert image into frequency
    syn_fft = fft(syn_img)
    real_fft = fft(rea_iImg)
    syn_amp, syn_phase = abs(syn_fft), angle(syn_fft)
    real_amp = abs(real_fft)
    syn_amp, real_amp = fftshift(syn_amp), fftshift(real_amp)

    # amplitude spectrum mixup
    aug_amp = G * syn_amp + (1 - G) * real_amp
    aug_amp = ifftshift(aug_amp)
    aug_fft = aug_amp * exp(1j * syn_phase)
    augImg = ifft(aug_fft)
    return augImg
```

Table 3: Notations and settings for SA about base augmentation and grayscale.

| Name | Description | $p^{LR}$ | $p^{Crop}$ | $p^{Pho}$ | $p^{Gray}$ |
|---|---|---|---|---|---|
| SA-B0 | Original in [4], base augmentation | 0.2 | 0.2 | 0.2 | - |
| SA-B1 | Stronger base augmentation | 0.5 | 0.5 | 0.5 | - |
| SA-B0-G0 | Add grayscale into SA-B0 | 0.2 | 0.2 | 0.2 | 0.01 |
| SA-B0-G1 | Add stronger grayscale into SA-B0 | 0.2 | 0.2 | 0.2 | 0.2 |
| SA-B1-G1 | Add stronger grayscale into SA-B1 | 0.5 | 0.5 | 0.5 | 0.2 |
| SA-B1-G2 | Add much stronger grayscale into SA-B1 | 0.5 | 0.5 | 0.5 | 0.4 |

the results are presented in Table 14. Regardless of the base augmentation, the random perspective consistently outperformed the random affine. Therefore, we have decided to include the random perspective in our data augmentation pipeline with a probability of 0.4.

### 3.3. Revisiting grayscale and similar augmentation

After selecting the basic appearance and geometry data augmentation, we find that appearance augmentation could efficiently improve the model generalization, leading to better performance in real-world image tests. To explore the potential of appearance augmentation while avoiding the risk of losing facial details, such as adding excessive noise or using GAN-based image editing that requires an additional model, we consider two smooth data augmentations: histogram equalization and autocontrast. As shown in the experiment details listed in Table 8 and the results in Table 7, both histogram equalization and autocontrast improve the test results, but histogram equalization has a greater impact than autocontrast. To keep the data augmentation list

Table 4: Notations and settings for SA about channel shuffle.

| Name | Description | $p^{LR}$ | $p^{Crop}$ | $p^{Pho}$ | $p^{Gray}$ | $p^{CS}$ |
|---|---|---|---|---|---|---|
| SA-B0-CS0 | Add Channel shuffle into SA-B0 | 0.2 | 0.2 | 0.2 | - | 0.01 |
| SA-B1-CS1 | Add Channel shuffle into SA-B1 | 0.5 | 0.5 | 0.5 | - | 0.2 |
| SA-B2-CS1 | Add Channel shuffle into SA-B1-G1 | 0.5 | 0.5 | 0.5 | 0.2 | 0.2 |

Table 5: Notations and settings for SA about Gaussian noise and Gaussian blur.

| Name | Description | $p^{LR}$ | $p^{Crop}$ | $p^{Pho}$ | $p^{Gray}$ | $p^{GB}$ | $p^{GN}$ |
|---|---|---|---|---|---|---|---|
| SA-B2-GB0 | Add GN into SA-B1-G1 | 0.5 | 0.5 | 0.5 | 0.2 | 0.2 | - |
| SA-B2-GN0 | Add GN into SA-B1-G1 | 0.5 | 0.5 | 0.5 | 0.2 | - | 0.2 |
| SA-B2-GB0-GN0 | Add GN into SA-B1-G1 | 0.5 | 0.5 | 0.5 | 0.2 | 0.2 | 0.2 |
| SA-B3-GB1-GN1 | Add GN into SA-B1-G2 | 0.5 | 0.5 | 0.5 | 0.4 | 0.4 | 0.4 |

Table 6: Notations and settings for geometry methods.

| Name | Base | $p^{Per}$ | $p^{Aff}$ |
|---|---|---|---|
| SA-B1-Per0 | SA-B1 | 0.2 | - |
| SA-B1-Aff0 | SA-B1 | - | 0.2 |
| SA-B4-Per1 | SA-B3-GB1-GN1 | 0.4 | - |
| SA-B4-Aff1 | SA-B3-GB1-GN1 | - | 0.4 |

Table 7: Ablation study for histogram equalization and autocontrast.

| Data Aug. | LFW | AgeDB-30 | CA-LFW | CFP-FP | CP-LFW | Avg. |
|---|---|---|---|---|---|---|
| SA-B0-HE0 | 89.37 | 70.03 | 73.45 | 77.66 | 70.13 | 76.13 |
| SA-B0-HE1 | 90.77 | 72.85 | 73.93 | 79.17 | 70.68 | 77.48 |
| SA-B1-HE1 | 90.63 | 72.63 | 75.93 | 79.64 | 71.90 | 78.15 |
| SA-B1-HE2 | 91.00 | 74.10 | 76.10 | 79.93 | 72.17 | 78.66 |
| SA-B0-HE3 | **92.40** | 74.37 | 75.58 | 80.64 | 72.33 | 79.07 |
| SA-B1-HE3 | 92.17 | **75.07** | **77.08** | **81.46** | **74.23** | **80.00** |
| SA-B0-AC0 | 88.33 | 67.46 | 62.66 | 76.51 | 68.37 | 74.67 |
| SA-B0-AC1 | 88.70 | 70.22 | 72.65 | 77.09 | 69.88 | 75.71 |
| SA-B1-AC1 | 88.83 | 71.77 | 73.95 | 78.14 | 69.48 | 76.44 |
| SA-B1-AC2 | 89.48 | 71.60 | 73.02 | 78.29 | 70.78 | 76.63 |
| SA-B0-AC3 | 89.80 | 68.90 | 72.32 | 77.59 | 69.43 | 75.61 |
| SA-B1-AC3 | 89.35 | 72.63 | 73.87 | 79.11 | 71.28 | 77.25 |

Table 8: Notations and settings for histogram equalization and autocontrast.

| Name | Base | $p^{Equal}$ | $p^{AC}$ |
|---|---|---|---|
| SA-B0-HE0 | SA-B0 | 0.05 | - |
| SA-B0-HE1 | SA-B0 | 0.2 | - |
| SA-B1-HE1 | SA-B0 | 0.2 | - |
| SA-B1-HE2 | SA-B0 | 0.4 | - |
| SA-B0-HE3 | SA-B0 | 1.0 | - |
| SA-B1-HE3 | SA-B0 | 1.0 | - |
| SA-B0-AC0 | SA-B0 | - | 0.05 |
| SA-B0-AC1 | SA-B0 | - | 0.2 |
| SA-B1-AC1 | SA-B0 | - | 0.2 |
| SA-B1-AC2 | SA-B0 | - | 0.4 |
| SA-B0-AC3 | SA-B0 | - | 1.0 |
| SA-B1-AC3 | SA-B0 | - | 1.0 |

Table 9: Notations and settings for histogram equalization and grayscale with probability of 1.0.

| Name | Base | $p^{Gray}$ | $p^{Equal}$ |
|---|---|---|---|
| SA-B5-HE3 | SA-B4-Per1 | - | 1.0 |
| SA-B5-G3-HE3 | SA-B4-Per1 | 1.0 | 1.0 |

manageable, we decide to use only histogram equalization. In addition, we make an interesting discovery that the experiments of both histogram equalization and autocontrast suggest setting their probability to 1, which is counterintuitive.

We further conduct experiments on histogram equalization based on **SA-B4-Per1** in Section 3.2, but varied the probability of grayscale, as shown in Table 9. The results in Table 15 show that both grayscale and histogram equalization benefit from setting the probability to 1.0. Although **SA-B5-G3-HE3** does not have the highest average accu-

racy, we still chose it as our data augmentation combination to simplify the conclusion. The discussion for setting the probability to 1 is left as our future work.

### 3.4. Order of SA

After finalizing our SA list, **SA-B5-G3-HE3**, we consider the order of applying SA to further optimize our model performance. We focus on the order of grayscale and histogram equalization in combination with other SA methods. Moreover, we explore how many times we can apply

Table 10: Notations and settings for order. "T" denotes true, and "F" denotes false.

| Name | Base | $Gray_{first}$ | $Gray_{last}$ | $HE_{first}$ | $HE_{last}$ |
|---|---|---|---|---|---|
| SA-B6-O1 | SA-B5-G3-HE3 | T | F | F | T |
| SA-B6-O2 | SA-B5-G3-HE3 | F | T | F | T |
| SA-B6-O3 | SA-B5-G3-HE3 | F | T | T | F |
| SA-B6-O4 | SA-B5-G3-HE3 | T | T | T | F |
| SA-B6-O5 | SA-B5-G3-HE3 | F | T | T | T |
| SA-B6-O6 | SA-B5-G3-HE3 | T | F | T | T |
| SA-B6-O7 | SA-B5-G3-HE3 | T | T | F | T |
| SA-B6-O8 | SA-B5-G3-HE3 | T | T | T | T |

Table 11: Ablation study for base augmentation and grayscale.

| Data Aug. | LFW | AgeDB-30 | CA-LFW | CFP-FP | CP-LFW | Avg. |
|---|---|---|---|---|---|---|
| SA-B0 | 88.43 | 67.27 | 71.52 | 76.57 | 69.45 | 74.65 |
| SA-B1 | 88.77 | 69.40 | 72.05 | 76.50 | 70.50 | 75.44 |
| SA-B0-G0 | 88.88 | 71.30 | 73.15 | 78.10 | 70.85 | 76.45 |
| SA-B0-G1 | 90.33 | 72.10 | 75.43 | 79.53 | 72.05 | 77.89 |
| SA-B1-G1 | 91.55 | **76.72** | 77.12 | **81.27** | 73.05 | **79.94** |
| SA-B1-G2 | **91.83** | 75.23 | **77.38** | 81.09 | **73.67** | 79.84 |

Table 12: Ablation study for channel shuffle.

| Data Aug. | LFW | AgeDB-30 | CA-LFW | CFP-FP | CP-LFW | Avg. |
|---|---|---|---|---|---|---|
| SA-B0-CS0 | 87.97 | 70.13 | 72.82 | 77.43 | 68.87 | 75.44 |
| SA-B1-CS1 | 89.37 | 73.00 | 75.72 | 78.61 | 71.43 | 77.63 |
| SA-B2-CS1 | 91.87 | 75.45 | 77.47 | 81.09 | 73.52 | 79.88 |

Table 13: Ablation study for Gaussian noise and Gaussian blur.

| Data Aug. | LFW | AgeDB-30 | CA-LFW | CFP-FP | CP-LFW | Avg. |
|---|---|---|---|---|---|---|
| SA-B2-GB0 | 91.97 | 76.02 | 78.68 | 80.77 | 74.05 | 80.30 |
| SA-B2-GN0 | 91.45 | 74.85 | 76.75 | 81.86 | 72.67 | 79.51 |
| SA-B2-GB0-GN0 | 92.10 | 76.63 | 78.17 | 81.79 | 73.17 | 80.37 |
| SA-B3-GB1-GN1 | **94.00** | **77.92** | **79.68** | **82.39** | **77.05** | **82.21** |

Table 14: Ablation study for geometry methods.

| Data Aug. | LFW | AgeDB-30 | CA-LFW | CFP-FP | CP-LFW | Avg. |
|---|---|---|---|---|---|---|
| SA-B1-Per0 | 88.97 | 67.22 | 72.15 | 79.17 | 71.27 | 75.75 |
| SA-B1-Aff0 | 89.47 | 67.88 | 71.70 | 77.80 | 70.30 | 75.43 |
| SA-B4-Per1 | **94.00** | **77.75** | **79.97** | **84.17** | **78.38** | **82.85** |
| SA-B4-Aff1 | 93.98 | 77.15 | 77.63 | 79.81 | 74.13 | 80.54 |

Table 15: Experimental results of histogram equalization and grayscale with probability of 1.0.

| Data Aug. | LFW | AgeDB-30 | CA-LFW | CFP-FP | CP-LFW | Avg. |
|---|---|---|---|---|---|---|
| SA-B5-HE3 | 93.73 | 77.55 | 77.73 | 62.43 | 76.50 | 77.59 |
| SA-B5-G3-HE3 | **95.47** | **77.97** | **81.65** | **84.89** | **79.22** | **83.84** |

Table 16: Ablation study for SA order.

| Data Aug. | LFW | AgeDB-30 | CA-LFW | CFP-FP | CP-LFW | Avg. |
|---|---|---|---|---|---|---|
| SA-B6-O1 | 95.35 | 77.57 | 81.33 | 84.19 | 78.63 | 83.41 |
| SA-B6-O2 | 95.57 | 77.90 | **81.60** | 84.76 | 79.12 | 83.79 |
| SA-B6-O3 | 95.62 | 77.62 | 80.92 | **85.27** | **79.67** | 83.82 |
| SA-B6-O4 | 95.32 | 78.37 | 81.37 | 85.21 | 79.53 | **83.96** |
| SA-B6-O5 | 95.15 | 78.48 | 81.45 | 85.01 | 79.22 | 83.86 |
| SA-B6-O6 | 91.27 | 74.87 | 76.80 | 80.03 | 73.55 | 79.30 |
| SA-B6-O7 | 95.27 | **79.28** | 81.43 | 84.50 | 79.03 | 83.90 |
| SA-B6-O8 | **95.72** | 78.73 | 81.15 | 84.57 | 79.42 | 83.92 |

before other data augmentation methods. We test different orders of applying SA, as shown in Table. 16. We use the notation $Gray_{first}$ to indicate applying grayscale first, and $HE_{first}$ to indicate applying histogram equalization first. If both adopted simultaneously, we apply grayscale first followed by histogram equalization. $Gray_{last}$ and $HE_{last}$ denote applying at the last. We choose **SA-B6-O4** as our final proposed SA based on the average accuracy.

## 4. Low-frequency vs high-frequency

In this study, the third assumption of SMU method mentioned in Section 3.3 is that enhancing high-frequency information is more effective than low-frequency information for domain generalization. However, previous mixup methods, such as [14, 5], immigrate low-frequency components from real image to synthetic image, rather than high-frequency ones. Hence, we conduct the experiments to verify our assumption, in which we swap the masks used to combine real amplitude spectrum and synthetic amplitude spectrum in [14, 5] methods. As shown in Table 17, their performance can be improved (please see the accuracy with green color) by modifying with our assumption. Moreover, considering the importance of different frequencies can obtain better performance than the methods which swap or weighted sum for synthetic and real amplitude spectra [13, 12].

Furthermore, we visualize the augmented images using different methods, as shown in Fig. 1. The images with red boxes are generated by those methods with our assumption, their PSNR values are higher than the original methods.

## 5. Conclusion

In this supplemetary meterial, we first demonstrate our proposed method is generalized to other dataset, which outperform other state-of-the-art methods with the same condition. Next, we provide implementation details of both SA and SMU. We then report on the step-by-step selection process of SA and provide extensive ablation studies to support our choice. Finally, we analyze the impact of low-frequency and high-frequency in frequency mixup methods and verify our assumption.

grayscale or histogram equalization. Our initial implementation applied both grayscale and histogram equalization

Table 17: Comparison of different mixup methods in the frequency domain (%). Note that previous mixup methods, such as [14, 5], are modified with our assumption that enhancing high-frequency information is more effective than low-frequency information. The accuracy in black color is that immigrates high-frequency information from real data to synthetic data, and the accuracy in green color are improvements compared with immigrating low-frequency ones.

| Method | LFW | AgeDB-30 | CA-LFW | CFP-FP | CP-LFW | Avg |
|---|---|---|---|---|---|---|
| - | 88.43 | 67.27 | 71.52 | 76.57 | 69.45 | 74.65 |
| Yang et al. [13] | 91.92 | 68.85 | 76.93 | 78.39 | 71.88 | 77.59 |
| Yang et al. [14] | 91.53 +5.08 | 71.22 +2.89 | 74.97 +3.60 | 79.37 +4.97 | 73.7 +6.33 | 78.16 +4.58 |
| Xu et al. [12] | 91.18 | 68.72 | 75.48 | 78.06 | 72.30 | 77.15 |
| Liu et al. [5] | 91.08 +4.06 | 71.28 +3.26 | 75.30 +3.63 | 79.36 +2.79 | 71.25 +3.82 | 77.65 +3.51 |
| Ours | **91.23** +2.21 | **73.62** +5.27 | **76.45** +3.12 | **80.43** +6.37 | **74.08** +5.40 | **79.16** +4.45 |

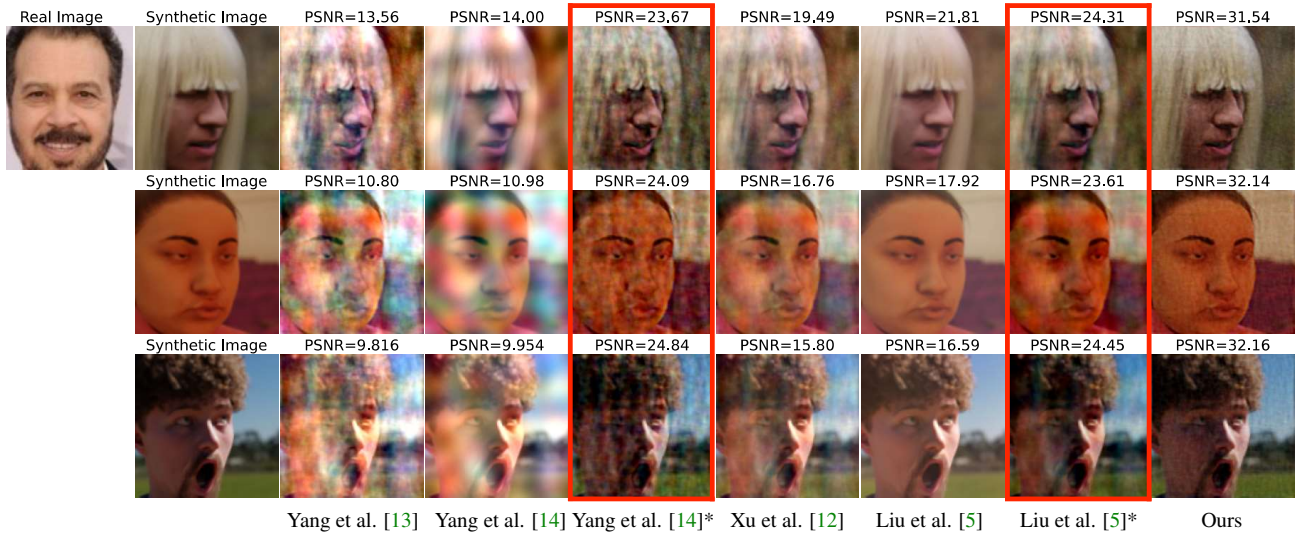

Figure 1: The visualization results using different mixup strategies in the frequency domain and PSNR values indicating the image quality and similarity between the original synthetic image and the augmented image. "*" means that we modify their methods with our assumption, in which we immigrate high-frequency information from real image to synthetic image.

## References

[1] Gwangbin Bae, Martin de La Gorce, Tadas Baltrušaitis, Charlie Hewitt, Dong Chen, Julien Valentin, Roberto Cipolla, and Jingjing Shen. Digiface-1m: 1 million digital face images for face recognition. In *Proceedings of the IEEE/CVF Winter Conference on Applications of Computer Vision*, pages 3526–3535, 2023. 1

[2] Fadi Boutros, Marcel Klemt, Meiling Fang, Arjan Kuijper, and Naser Damer. Unsupervised face recognition using unlabeled synthetic data. In *2023 IEEE 17th International Conference on Automatic Face and Gesture Recognition (FG)*, pages 1–8. IEEE, 2023. 1, 2

[3] Yu Deng, Jiaolong Yang, Dong Chen, Fang Wen, and Xin Tong. Disentangled and controllable face image generation via 3d imitative-contrastive learning. In *Proceedings of the IEEE/CVF conference on computer vision and pattern recognition*, pages 5154–5163, 2020. 1

[4] Minchul Kim, Anil K Jain, and Xiaoming Liu. Adaface: Quality adaptive margin for face recognition. In *Proceedings of the IEEE/CVF Conference on Computer Vision and Pattern Recognition*, pages 18750–18759, 2022. 1, 2, 3

[5] Quande Liu, Cheng Chen, Jing Qin, Qi Dou, and Pheng-Ann Heng. Feddg: Federated domain generalization on medical image segmentation via episodic learning in continuous frequency space. In *Proceedings of the IEEE/CVF Conference on Computer Vision and Pattern Recognition*, pages 1013–1023, 2021. 5, 6

[6] TorchVision maintainers and contributors. Torchvision: Pytorch's computer vision library. https://github.com/pytorch/vision, 2016. 2

[7] Alhassan Mumuni and Fuseini Mumuni. Data augmentation: A comprehensive survey of modern approaches. *Array*, page 100258, 2022. 1

[8] Henri J Nussbaumer and Henri J Nussbaumer. *The fast Fourier transform*. Springer, 1982. 1

[9] Haibo Qiu, Baosheng Yu, Dihong Gong, Zhifeng Li, Wei Liu, and Dacheng Tao. Synface: Face recognition with syn-

thetic data. In *Proceedings of the IEEE/CVF International Conference on Computer Vision*, pages 10880–10890, 2021. 2

[10] Connor Shorten and Taghi M Khoshgoftaar. A survey on image data augmentation for deep learning. *Journal of big data*, 6(1):1–48, 2019. 1

[11] Stefan Van der Walt, Johannes L Schönberger, Juan Nunez-Iglesias, François Boulogne, Joshua D Warner, Neil Yager, Emmanuelle Gouillart, and Tony Yu. scikit-image: image processing in python. *PeerJ*, 2:e453, 2014. 2

[12] Qinwei Xu, Ruipeng Zhang, Ya Zhang, Yanfeng Wang, and Qi Tian. A fourier-based framework for domain generalization. In *Proceedings of the IEEE/CVF Conference on Computer Vision and Pattern Recognition*, pages 14383–14392, 2021. 5, 6

[13] Yanchao Yang, Dong Lao, Ganesh Sundaramoorthi, and Stefano Soatto. Phase consistent ecological domain adaptation. In *Proceedings of the IEEE/CVF Conference on Computer Vision and Pattern Recognition*, pages 9011–9020, 2020. 5, 6

[14] Yanchao Yang and Stefano Soatto. Fda: Fourier domain adaptation for semantic segmentation. In *Proceedings of the IEEE/CVF Conference on Computer Vision and Pattern Recognition*, pages 4085–4095, 2020. 5, 6



\end{document}